\documentclass{article} %
\usepackage{conference_style,times}
\conffinalcopy

\usepackage{hyperref}
\usepackage{url}
\hypersetup{hidelinks,pdftitle={Hierarchical Continuous Diffusion Language Models}}
\hypersetup{pdfauthor={Hui Ren, Zihan Li, Chang Liu, Huidong Liu, Alexander Schwing}}
\usepackage{amsmath,amssymb,amsthm}
\usepackage{mathtools}
\usepackage{nicefrac}
\usepackage{microtype}
\allowdisplaybreaks

\usepackage{graphicx}
\usepackage{xcolor}
\usepackage{booktabs}
\usepackage{multirow}
\usepackage{makecell}
\usepackage{subcaption}
\usepackage{wrapfig}
\newcommand{\repwidth}{1.0}\newcommand{\subfigureseparator}{\par\vspace{3mm}}
\newcommand{\objectivelinebreak}{\\*[6pt]}
\usepackage{algorithm}
\usepackage{algpseudocode}

\usepackage[capitalize]{cleveref}

\newtheorem{proposition}{Proposition}

\newtheorem{remark}{Remark}

\DeclareMathOperator*{\argmax}{arg\,max}
\newcommand{\E}{\mathbb{E}}
\newcommand{\KL}{\mathrm{KL}}
\newcommand{\calL}{\mathcal{L}}
\newcommand{\calJ}{\mathcal{J}}

\definecolor{linkcolor}{RGB}{18,88,138}
\newcommand{\name}{HC-DLM}
\newcommand{\fullname}{Hierarchical Continuous Diffusion Language Models}

\title{\fullname}

\newcommand{\projectpage}{https://hc-dlm.github.io/}
\author{Hui Ren$^{1}$ \quad Zihan Li$^{1}$ \quad Chang Liu$^{1}$ \quad Huidong Liu$^{2}$ \quad Alexander Schwing$^{1}$ \\
$^{1}$University of Illinois Urbana-Champaign \qquad $^{2}$Amazon.com, Inc.}

\begin{document}

\maketitle
\lhead{}
\begin{abstract}
Discrete diffusion language models offer a compelling alternative to autoregressive generation for tasks demanding bidirectional reasoning and global constraint satisfaction. Yet they share a structural bottleneck: when decoding in parallel, each token is sampled independently from its marginal, severing the statistical dependencies among the tokens decoded together. %
Continuous diffusion language models avoid this by denoising a shared continuous state, but their denoiser sees only that state, so nothing ties  it to a valid token configuration until it is finally decoded.
To address this, we propose \textbf{\fullname{} (\name{})}, which couple discrete token generation with a continuous latent trajectory in a single, principled denoising process, whose training objective is derived from a variational bound on the token likelihood. In contrast to recent methods that attach continuous context to a self-contained discrete chain, \name{} makes the latent the only persistent generative state: tokens are read out from it at every step and feed back as a scaffold for the next latent update. %
On structured reasoning (Sudoku), mathematical planning (Countdown) and language modeling (LM1B), \name{} improves over discrete and continuous diffusion baselines at matched model size, in puzzle accuracy on Sudoku and Countdown and in generative perplexity on LM1B. \ \textbf{Project page:} \url{\projectpage}. %

\end{abstract}

\section{Introduction}

Autoregressive (AR) language models generate text left-to-right. %
This sequential factorization is remarkably successful, %
yet it is a mismatch for tasks that need global constraint satisfaction or bidirectional computation, such as solving logical puzzles or planning arithmetic operations. %
On these tasks, AR models commit early to choices that are challenging to revise~\citep{ye2024beyond}. %

Discrete diffusion language models~\citep{austin2021structured, lou2023discrete, sahoo2024simple, nie2025large} provide a principled alternative. By modeling the joint distribution over all tokens via an iterative denoising process, they enable the model to condition on arbitrary subsets of tokens and refine its predictions over many bidirectional passes, decoding multiple tokens in parallel at each step. This yields empirical gains on reasoning and planning~\citep{ye2024beyond, kim2025train} and, at scale, perplexity competitive with strong AR baselines~\citep{nie2025large, lou2023discrete}. %

However, a bottleneck remains at the heart of parallel decoding: \emph{token dependence}.
When the model unmasks multiple positions in a single denoising step, each position is independently sampled from its marginal conditioned on a partial sequence. Hence, the joint over the simultaneously decoded tokens is modeled as a product of marginals~\citep{gu2018nonautoregressive, li2026breaking, hersche2026locally, kim2026dependency, ringel2026dependency}, which fails precisely where tokens are tightly coupled by syntax, logic, or physical constraints.

Continuous diffusion language models address this by denoising a continuous state shared by all tokens: tokens decoded in parallel all depend on the same latent instead of being sampled on their own. They denoise either the token embeddings~\citep{li2022diffusion, gulrajani2023likelihood, chen2026langflow} or a compressed latent produced by an encoder~\citep{zhang2023planner, lovelace2023latent, guo2026continuous}, and decode tokens from the clean result. Their weakness mirrors the discrete one. Their denoiser takes only the continuous state as input, and no term of the objective involves tokens along the trajectory, removing ties to a valid token configuration until the final decoding step. Rounding~\citep{li2022diffusion} and self-conditioning~\citep{chen2022analog, chen2026langflow} feed the model's own token estimate back as an input, but they share the same problem, since that estimate is neither a variable of the model nor a term of its objective.

\begin{figure}[t]
  \centering
  \includegraphics[width=1.0\linewidth]{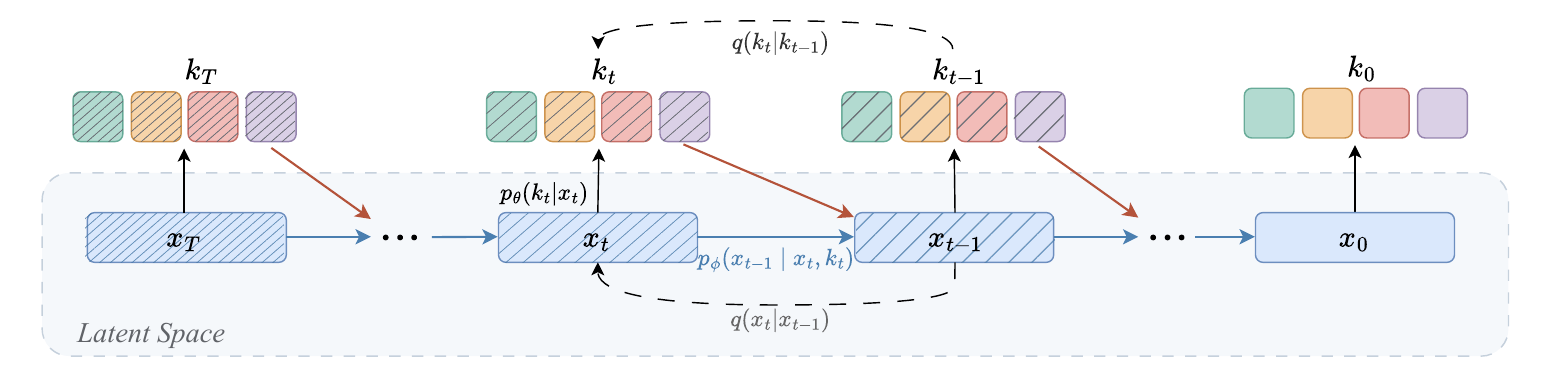}
  \vspace{-3mm}
  \caption{%
    \textbf{\name{} coupled diffusion process.}
    The forward direction independently corrupts the continuous latent trajectory
    $x_0 \to x_T$ and the discrete token trajectory $k_0 \to k_T$ via
    $q(x_t \mid x_{t-1})$ and $q(k_t \mid k_{t-1})$.
    The reverse direction couples the two channels: $p_\phi(x_{t-1}\mid x_t,k_t)$
    advances the continuous state conditioned on the current token state, while
    $p_\theta(k_t\mid x_t)$ reads token distributions from the latent state, iterating until the clean states $(x_0,k_0)$ are reached.
  }
  \label{fig:coupled_process}
  \vspace{-3mm}
\end{figure}

The two weaknesses are complementary, and so are their remedies. Tokens decoded in parallel need a shared state to depend on, which a continuous latent provides, and a latent being denoised needs a discrete configuration to be tied to, which the tokens provide. We therefore propose \textbf{\fullname{} (\name{})}, a generative framework whose reverse process is a two-level chain (\cref{fig:coupled_process}). The continuous latent trajectory $\{x_t\}$ is the only state that persists across steps. At every reverse step, tokens are read out from the current latent through a readout distribution $p_\theta(k_t\mid x_t)$, re-noised through the discrete forward kernel, and fed back as the condition of the next latent transition $p_\phi(x_{t-1}\mid x_t,k_t)$. The token state is thus a variable of the generative model with a forward kernel of its own, which is what separates the scaffold from a fed-back estimate. In the generative model, $k_t$ depends on the past only through $x_t$, so the tokens carry no transition chain of their own and every position is read out anew at each step. We call this organization the hierarchical coupling. It determines both how the model is trained and how it samples: a single variational bound over the two-level trajectory gives the training objective (\cref{sec:elbo}), and generation alternates latent denoising with token readout and re-noising (\cref{sec:inference}).

Hybrid discrete--continuous diffusion models also pair the two spaces, through per-token continuous hints (CADD)~\citep{zheng2025continuously}, or an embedding chain denoised in parallel with the tokens (CCDD)~\citep{zhou2025coevolutionary}. However, separate transition chains are used as discussed in %
\cref{sec:related} and \cref{app:comparison}. %

\noindent\textbf{Contributions.}
\vspace{-3mm}
\begin{enumerate}
    \setlength\itemsep{0pt}
    \item \textbf{Framework.}
    We introduce \name{} (\cref{sec:model}), a hierarchical generative structure for discrete sequences, in which a continuous latent is the only persistent state and tokens are per-step readouts that feed back as a conditioning scaffold. 

    \item \textbf{A likelihood bound for the hierarchical coupling chain.}
    We derive a variational lower bound for \name{} (\cref{sec:elbo}) over the hierarchical trajectory,  which establishes that reading tokens from the latent and feeding them back is a proper generative model of the token sequence.

    \item \textbf{Both the continuous latent and the token feedback are needed.}
    We show on Sudoku, Countdown and LM1B that \name{} improves over discrete and continuous diffusion baselines at matched model size (\cref{sec:experiments}), and that removing either one leaves a model that falls well short of the full one (\cref{sec:ablation}).
\end{enumerate}
\vspace{-2mm}

\section{Preliminaries}
\label{sec:prelim}

\noindent\textbf{Discrete Diffusion Models.}
A discrete diffusion language model~\citep{austin2021structured, sahoo2024simple} defines a forward Markov chain that progressively corrupts a discrete length $L$ token sequence $k_0 = (\dots, k_0^i, \dots) \in \mathcal{V}^L$  over vocabulary ${\cal V}$ ($|\mathcal{V}| = K$) with $i\in\{1, \dots, L\}$. Using a noise schedule $\alpha_0 = 1 \geq \alpha_1 \geq \cdots \geq \alpha_T \approx 0$, the marginal at time $t\in\{0, \dots, T\}$ interpolates between sequence $k_0$ and stationary distribution $\pi$:
\begin{equation}
    q(k_t^i \mid k_0^i)
    = \alpha_t\,\delta(k_t^i - k_0^i)
    + (1 - \alpha_t)\,\pi(k_t^i),
    \quad
    q(k_t^i \mid k_{t-1}^i)
    = \frac{\alpha_t}{\alpha_{t-1}}\,\delta(k_t^i - k_{t-1}^i)
    + \Bigl(1 - \frac{\alpha_t}{\alpha_{t-1}}\Bigr)\,\pi(k_t^i).
    \label{eq:discrete_forward}
\end{equation}
Following~\citet{austin2021structured}, two common choices of $\pi$ recover the standard discrete diffusion formulations:
\emph{(i) absorbing (\textsc{mask}) state}, $\pi(k) = \delta(k - \textsc{[mask]})$,
which corrupts each token toward a special mask symbol and underlies most masked diffusion language models~\citep{sahoo2024simple,nie2025large};
\emph{(ii) uniform state}, $\pi(k) = 1/K$,
which replaces tokens by a uniform draw over the vocabulary. A parametric model $p_\theta(k_0 \mid k_t)$, typically a bidirectional transformer trained with token-level cross-entropy, reverses this corruption.

\noindent\emph{Token independence.} 
During inference, the factored form $p_\theta(k_0 \mid k_t) = \prod_i p_\theta(k_0^i \mid k_t)$ samples each token independently. While the shared context $k_t$ provides some global information, the joint distribution over simultaneously decoded tokens is modeled as a product of marginals which does not capture their statistical dependencies.

\noindent\textbf{Continuous Diffusion Models.}
For continuous data $x_0 \in \mathbb{R}^d$, a DDPM~\citep{ho2020denoising} defines a Gaussian forward process with tractable posterior $q(x_{t-1} \mid x_t, x_0)$ and trains a denoiser via a weighted MSE objective. Flow Matching (FM)~\citep{lipman2022flow, liu2022flow} takes a more direct approach, training a velocity network $v_\phi$ to regress a target vector field transporting noise $\epsilon\sim\mathcal{N}(0,I)$ to data $x_0$. Using the affine interpolation $x_t = (1-t)\,x_0 + t\,\epsilon$, the Conditional Flow Matching objective is:
$    \calL_{\text{CFM}}
    = \E_{t,\;x_0,\;\epsilon}\bigl[
        \|v_\phi(x_t,\,t) - (\epsilon - x_0)\|^2
      \bigr].
$
Samples are generated by integrating $\dot{x} = v_\phi(x, t)$ backward from $t=1$ to $t=0$, often with  fewer function evaluations than DDPM sampling.

\section{\fullname}
\label{sec:method}

In this section, we first define a coupled continuous--discrete trajectory whose forward corruption remains tractable while its reverse dynamics are organized into two levels, with tokens read out from the latent state and fed back as a scaffold for the next latent update. We then derive a variational lower bound for this hierarchical process, convert it into a practical training objective (\cref{fig:train}), and describe the resulting alternating sampler.

\subsection{Generative Model and Forward Process}
\label{sec:model}

The token-independence bottleneck calls for a reverse process whose state carries cross-token information before any token is final, and the weakness of latent diffusion calls for that state to stay anchored to a token sequence while it is denoised. We therefore make the continuous representation itself a denoising trajectory and let every reverse step update both the continuous state and the token scaffold that conditions it.

For this, we model a sequence of discrete tokens $k_0 \in \mathcal{V}^L$ by introducing a coupled continuous latent state $x_0 \in \mathbb{R}^{M \times d}$. $M$ is the latent sequence length and $d$ is the per-position embedding dimension. The model maintains two trajectories, a continuous one $\{x_t\}_{t=0}^{T}$ and a discrete one $\{k_t\}_{t=0}^{T}$, that are corrupted independently in the forward direction but tightly coupled in the reverse direction.

\noindent\textbf{Variational distribution (forward process).}
For independent noising, we use %
\begin{equation}
    q_\psi(x_{0:T},\, k_{1:T} \mid k_0)
    =
    q_\psi(x_0 \mid k_0)
    \prod_{t=1}^T
    q(x_t \mid x_{t-1})\;
    q(k_t \mid k_{t-1}).
    \label{eq:forward}
\end{equation}
The only learned component is the \emph{encoder} $q_\psi(x_0\mid k_0)$, which produces a continuous latent representation of the clean token sequence. Jointly training it with the generative model permits the latent geometry to adapt to the needs of denoising. The continuous noising $q(x_t\mid x_{t-1})$ is a standard Gaussian diffusion process with tractable marginals $q(x_t\mid x_0)$ and closed-form posteriors $q(x_{t-1}\mid x_t,x_0)$. It converges to a standard Gaussian prior $q(x_T\mid x_0)\!\approx\!\mathcal{N}(0,I)$ at the terminal time. %
For the discrete noising kernel $q(k_t\mid k_{t-1})$ we use the categorical forward process defined in  \cref{eq:discrete_forward}, with either an absorbing or uniform state distribution. Note that the two forward chains in \cref{eq:forward} are \emph{independent}: $x_t$ is corrupted by Gaussian noise while $k_t$ is corrupted separately by token-level noise. This independence keeps both forward marginals tractable, exactly as in standard continuous and discrete diffusion. %

\noindent\textbf{Generative model (reverse process).}
While the forward chains are independent, the reverse chain deliberately breaks this symmetry: a continuous denoiser $p_\phi(x_{t-1}\mid x_t,k_t)$ conditions on the current token state. Hence, information flows from  discrete variables into the continuous trajectory at every step. %
Specifically, the joint generative model factorizes as follows: 
\begin{equation}
    p_{\theta,\phi}(k_{0:T},\, x_{0:T})
    =
    p(x_T)\; p_\theta(k_0 \mid x_0)
    \prod_{t=1}^T
    p_\theta(k_t \mid x_t)\;
    p_\phi(x_{t-1} \mid x_t, k_t).
    \label{eq:gen_model}
\end{equation}
Two learned distributions drive the reverse direction of \cref{fig:coupled_process}. The \emph{token predictor} $p_\theta(k_t\mid x_t)$, with parameters $\theta$, maps the noisy continuous state to a distribution over tokens at time $t$: at $t{=}0$ it produces the final discrete output, while at intermediate $t$ it acts as a soft readout that maps the continuous state back to a distribution over the discrete vocabulary. The \emph{continuous denoiser} $p_\phi(x_{t-1}\mid x_t, k_t)$, with parameters $\phi$, advances the latent state conditioned on the current token state. Importantly, dependence of the denoising on $k_t$  makes \name{} more than two parallel diffusion processes: the token state acts as a discrete scaffold that resolves continuous-space ambiguity. Confident token estimates constrain the directions in which $x_t$ can move, while uncertain positions leave the continuous state free to explore alternative completions.

This intuition can be made precise. %
This reverse chain is a hierarchy rather than a pair of sibling processes, that is, two self-contained chains coupled at the same level. %
The token state has no transition kernel of its own: in \cref{eq:gen_model}, $k_t$ depends on the past only through $x_t$, so all cross-step information travels through the continuous trajectory, and tokens are read out anew at every step, which keeps them revisable.

\subsection{Variational Lower Bound}
\label{sec:elbo}

\begin{figure}[t]
  \centering
  \includegraphics[width=1.0\linewidth]{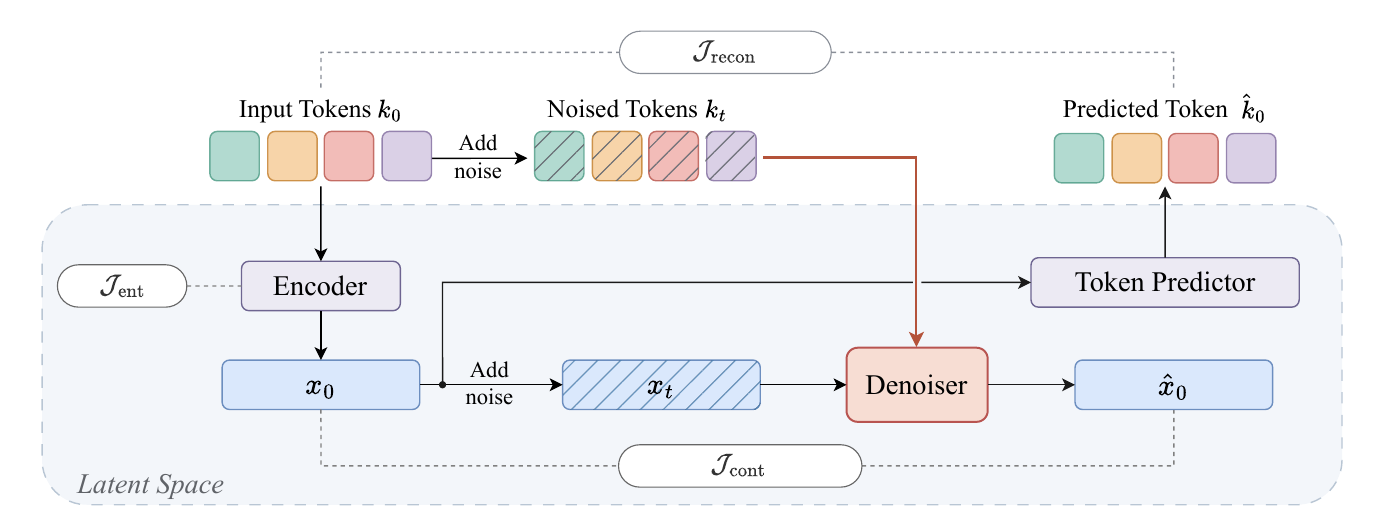}
  \vspace{-3mm}
  \caption{%
    \textbf{\name{} training pipeline.}
    The encoder $q_\psi(x_0\mid k_0)$ maps clean tokens $k_0$ to a continuous latent
    $x_0$, while independent forward kernels produce the noisy pair $(x_t,k_t)$.
    The token-conditioned denoiser $d_\phi(x_t,k_t,t/T)$ produces the clean-latent
    estimate $\hat{x}_0$ (loss $\calJ_{\mathrm{cont}}$), the token predictor
    $p_\theta(k_0\mid x_0)$ decodes $x_0$ to $\hat{k}_0$ (loss
    $\calJ_{\mathrm{recon}}$), and the encoder is regularized by
    $\calJ_{\mathrm{ent}}$.
  }
  \label{fig:train}
  \vspace{-3mm}
\end{figure}

The two-level reverse chain also needs a principled training objective, and the question is whether coupling the reverse dynamics keeps the forward process simple enough for variational learning. It does: because the continuous and discrete corruptions in \cref{eq:forward} are independent, the two processes above admit the following bound, which separates into interpretable reconstruction, prior-matching, boundary, and denoising terms.
\begin{proposition}[\name{} ELBO]
\label{prop:elbo}
Given the generative model \cref{eq:gen_model} and variational distribution \cref{eq:forward}, we obtain $\log p(k_0) \geq \calL(k_0)$ with
\begin{align}
\calL(k_0)
&= \underbrace{
    \E_{q_\psi(x_0 \mid k_0)}\bigl[\log p_\theta(k_0 \mid x_0)\bigr]
   }_{\text{(i) reconstruction}}
-\underbrace{
    \E_{q_\psi(x_0 \mid k_0)}\left[
      \KL\bigl(q(x_T \mid x_0)\,\|\,p(x_T)\bigr)
    \right]
   }_{\text{(ii) prior matching}}
\notag\\
&\quad+\;
\underbrace{
    \E_{q_\psi(x_0, x_1 \mid k_0),\,q(k_1 \mid k_0)}\!\bigl[\log p_\phi(x_0 \mid x_1, k_1)\bigr]
   }_{\text{(iii.a) boundary continuous denoising}}
\notag\\
&\quad+\;
\underbrace{
    H\!\bigl(q_\psi(x_0 \mid k_0)\bigr)
   }_{\text{(iii.b) encoder entropy}}
\;-\;
\underbrace{
    \E_{q_\psi(x_1 \mid k_0)}\!\bigl[
      \KL\bigl(q(k_1 \mid k_0)\,\|\,p_\theta(k_1 \mid x_1)\bigr)
    \bigr]
   }_{\text{(iii.c) boundary discrete token prediction}}
\notag\\
&\quad-
\underbrace{
    \sum_{t=2}^T
    \E_{q_\psi(x_0, x_t, k_{t-1} \mid k_0)}\!\bigl[
      \KL\bigl(
        q(x_{t-1} \mid x_t, x_0)q(k_t \mid k_{t-1})
        \|
        p_\phi(x_{t-1} \mid x_t, k_t)p_\theta(k_t \mid x_t)
      \bigr)
    \bigr]
   }_{\text{(iv) denoising step}}.
\label{eq:elbo}
\end{align}
\end{proposition}
The derivation is provided in \cref{app:proof}. We provide some intuition for each term next:
\textbf{(i)} a data reconstruction log-likelihood. %
\textbf{(ii)} a terminal prior-matching, which doesn't contain trainable parameters.
\textbf{(iii.a)} the boundary continuous denoising log-likelihood. %
\textbf{(iii.b)} an encoder entropy, which encourages uncertainty and hence a non-degenerate variational encoder.
\textbf{(iii.c)} the boundary discrete token prediction cross-entropy.
Finally, \textbf{(iv)} contributes two signals for every $t\ge2$, since the reverse model predicts a token state and a continuous state at every step.
Concretely, applying the chain rule of the KL divergence, and using the fact that $q(x_{t-1}\mid x_t,x_0)$ does not depend on $k_t$ because the continuous and discrete forward kernels are independent in \cref{eq:forward}, yields
\begin{align}
&\KL\Bigl(q(x_{t-1} \mid x_t, x_0)\,q(k_t \mid k_{t-1})\,\Big\|\,
            p_\phi(x_{t-1} \mid x_t, k_t)\,p_\theta(k_t \mid x_t)\Bigr)
\notag\\*
&=
\underbrace{\KL\bigl(q(k_t \mid k_{t-1})\,\|\,p_\theta(k_t \mid x_t)\bigr)}_{\text{discrete token prediction}}
+
\underbrace{
  \E_{q(k_t \mid k_{t-1})}\!\left[
    \KL\bigl(q(x_{t-1} \mid x_t, x_0)\,\|\,p_\phi(x_{t-1} \mid x_t, k_t)\bigr)
  \right]
}_{\text{continuous denoising}}.
\label{eq:kl_decompose}
\end{align}
The first sub-term is the discrete token prediction loss. %
The second sub-term is the continuous denoising loss, which we implement with flow matching rather than by fitting an explicit Gaussian reverse kernel. A detailed derivation of this identity is given in \cref{app:proof}.

\begin{remark}[Connection to discrete diffusion]
The discrete token prediction sub-term in \cref{eq:kl_decompose} has the same functional form as the token-prediction KL used in categorical discrete diffusion, with $x_t$ playing the role of the noisy conditioning state. \name{} therefore retains discrete-diffusion supervision while augmenting it with a continuous denoising signal. %
\end{remark}

\subsection{Practical Training Objective}
\label{sec:training_obj}

The ELBO identifies three trainable signals: reconstruct the clean tokens from the clean latent, keep the encoder distribution non-degenerate, and denoise the continuous latent trajectory conditioned on the current token. We write expectations under $q_\psi$ for samples from the forward/variational process.

Using the standard Gaussian parameterization of continuous diffusion, the continuous likelihood/KL terms reduce to a weighted MSE denoising objective~\citep{ho2020denoising,luo2022understand}, with schedule-dependent target $d_t(x_t,x_0)$ and weight $\omega_t$. The explicit derivation is provided in \cref{app:proof}.

To obtain a loss, two points are notable. First, the boundary continuous denoising term~\textbf{(iii.a)} has the same functional form as a $t{=}1$ instance of the continuous denoising sub-term in~\cref{eq:kl_decompose}, so we include it in the sum over timesteps. Second, we use an $x_0$-prediction parameterization for the token predictor: from the noisy state $x_t$ we form a clean-latent estimate $\hat{x}_0(x_t)$ via the denoiser $d_\phi$ (the explicit form depends on the parameterization of $d_\phi$, \cref{app:arch}), and the token predictor maps $\hat{x}_0(x_t)$ to a clean-token distribution. The corresponding noisy-token distribution is obtained by applying the known discrete forward kernel. Under this parameterization, the intermediate noisy-token KL (the discrete token prediction sub-term in \cref{eq:kl_decompose}) is controlled by a clean-token cross-entropy term; for the absorbing kernel it reduces to the same clean-token loss up to a schedule-dependent weight and constants, while for general categorical kernels it follows from an upper bound by the data processing inequality (\cref{app:proof}).
Hence, term~\textbf{(i)} and the boundary/intermediate discrete token prediction terms (from \textbf{(iii.c)} and \cref{eq:kl_decompose}) are represented by $\calJ_{\mathrm{recon}}$; the boundary continuous denoising term~\textbf{(iii.a)} and the continuous denoising sub-term of \cref{eq:kl_decompose} becomes $\calJ_{\mathrm{cont}}$; \textbf{(iii.b)} becomes $\calJ_{\mathrm{ent}}$.  %
Combined we have
\begin{align}
\calJ
&=
\lambda_{\mathrm{recon}}
\underbrace{
  \E_{q_\psi(x_0 \mid k_0)}\!\bigl[-\log p_\theta(k_0 \mid x_0)\bigr]
}_{\calJ_{\mathrm{recon}}:\;\text{reconstruction}}
\;+\;
\lambda_{\mathrm{ent}}
\underbrace{
  \E_{q_\psi(x_0 \mid k_0)}\!\bigl[\log q_\psi(x_0 \mid k_0)\bigr]
}_{\calJ_{\mathrm{ent}}:\;\text{encoder entropy}}
\notag\objectivelinebreak
&\quad+\;
\lambda_{\mathrm{cont}}
\underbrace{
  \sum_{t=1}^{T}
  \E_{q_\psi(x_0,x_t,k_t\mid k_0)}
  \!\left[
    \omega_t
    \bigl\|d_\phi(x_t,k_t,t/T)-d_t(x_t,x_0)\bigr\|^2
  \right]
}_{\calJ_{\mathrm{cont}}:\;\text{continuous denoising}}.
\label{eq:obj_simplified}
\end{align}
Here, $\calJ_{\mathrm{recon}}$ encourages that $x_0$ can be decoded to $k_0$, $\calJ_{\mathrm{cont}}$ trains the token-conditioned continuous denoising trajectory, and $\calJ_{\mathrm{ent}}$ prevents the Gaussian encoder from collapsing to a deterministic map~\citep{kingma2022autoencodingvariationalbayes,higgins2017beta}. In experiments, we instantiate $\calJ_{\mathrm{cont}}$ with Conditional Flow Matching (CFM). This implementation choice is detailed in \cref{app:proof}. $\calJ_{\mathrm{recon}}$ evaluates the predictor at the encoder sample $x_0$ rather than at $\hat{x}_0(x_t)$, so token-level gradients do not reach the denoiser. 
In practice we sample a single timestep per training example and form a Monte Carlo estimate of \cref{eq:obj_simplified}. \cref{alg:train} summarizes the inner loop.

\begin{algorithm}[t]
\caption{\name{} Training (single-sample Monte Carlo)}
\label{alg:train}
\begin{algorithmic}[1]
\Require Dataset $\mathcal{D}$; discrete noise schedule $\{\alpha_t\}$; loss weights $\lambda_{\mathrm{recon}}, \lambda_{\mathrm{cont}}, \lambda_{\mathrm{ent}}$
\Repeat
    \State Sample $k_0 \sim \mathcal{D}$, \; $t \sim \mathrm{Uniform}(1, T)$
    \State $x_0 \sim q_\psi(x_0 \mid k_0)$ \Comment{Encode: tokens $\to$ continuous latent}
    \State Sample $x_t \sim q(x_t\mid x_0)$ \Comment{Continuous noising}
    \State $k_t \sim q(k_t \mid k_0)$ \Comment{Forward discrete noising}
    \State Compute $\calJ$ following \cref{eq:obj_simplified}
    \State Update $\theta, \phi, \psi$ to minimize $\calJ$
\Until{stopping criterion}
\end{algorithmic}
\end{algorithm}

\subsection{Inference}
\label{sec:inference}

Inference instantiates the coupled reverse chain in \cref{eq:gen_model}, made explicit in \cref{alg:sample}: starting from $x_T\sim\mathcal{N}(0,I)$ and $k_T\sim\pi$, each step (i)~forms the scaffold-conditioned clean-latent estimate $\hat{x}_0$ with the denoiser and advances the continuous state to $x_{t-1}$; (ii)~reads clean tokens $\hat{k}_0$ out of $\hat{x}_0$ with the token predictor; and (iii)~re-noises $\hat{k}_0$ through the known forward kernel to obtain $k_{t-1}$. After the final step, we return $\hat{k}_0$ from $p_\theta(k_0\mid x_0)$. Under the absorbing kernel, step~(iii) can replace random re-masking with adaptive, confidence-based re-masking (\cref{sec:ablation}). Architecture and conditional-generation details are in \cref{app:arch,app:cond_gen}.

\begin{algorithm}[t]
\caption{\name{} Sampling}
\label{alg:sample}
\begin{algorithmic}[1]
\Require Denoiser $d_\phi$; token predictor $p_\theta$; discrete forward kernel $q(k_t\mid k_0)$ (\cref{eq:discrete_forward}); steps $T$
\State $x_T \sim \mathcal{N}(0, I)$;\quad $k_T \sim \pi$ \Comment{Stationary token distribution}
\For{$t = T, \dots, 1$}
    \State $\hat{x}_0 \gets d_\phi(x_t, k_t, t/T)$ \Comment{Scaffold-conditioned clean-latent estimate}
    \State $x_{t-1} \gets x_t - \tfrac{1}{t}\,(x_t - \hat{x}_0)$ \Comment{Euler step (linear schedule)}
    \State $\hat{k}_0 \gets \argmax_{k_0}\, p_\theta(k_0 \mid \hat{x}_0)$ \Comment{Read tokens out of the latent}
    \State $k_{t-1} \sim q(k_{t-1} \mid k_0{=}\hat{k}_0)$ \Comment{Re-noise readout to level $t{-}1$}
\EndFor
\State \Return $\argmax_{k_0}\, p_\theta(k_0 \mid x_0)$
\end{algorithmic}
\end{algorithm}

\section{Experiments}
\label{sec:experiments}
\vspace{-2mm}
We evaluate \name{} on three tasks targeting complementary capabilities: structured reasoning (Sudoku), mathematical planning (Countdown), and general language generation (LM1B), comparing against autoregressive, discrete diffusion, and continuous or hybrid diffusion baselines.

\subsection{Setup and Baselines}
\label{sec:exp_setup}

\begin{wraptable}{r}{0.5\textwidth}
\vspace{-4mm}
\centering
\caption{Sudoku puzzle-solving accuracy (\%$\uparrow$) on easy and hard splits. Best result per column in \textbf{bold}.}
\vspace{-2mm}
\label{tab:sudoku}
\setlength{\tabcolsep}{4pt}
{\small
\begin{tabular}{lrrr}
\toprule
Method & \#Params & Easy & Hard \\
\midrule
\multicolumn{4}{l}{\textit{Autoregressive}} \\
ARM (w/o ordering)  & \multirow{2}{*}{42M}  & 9.73  & -- \\
ARM (with ordering) &  & 87.18 & 32.57 \\
\midrule
\multicolumn{4}{l}{\textit{Discrete diffusion}} \\
MDM (vanilla)           & \multirow{3}{*}{6M} & 6.88  & 3.62 \\
MDM (top-prob.)         &                     & 18.51 & 9.44 \\
MDM (top-prob.\ margin) &                     & 89.49 & 49.88 \\
\midrule
\multicolumn{4}{l}{\textit{Hybrid discrete--continuous}} \\
CCDD & 6M & \textbf{94.65} & 70.73 \\
\midrule
\name{} (ours) & 6M & 94.21 & \textbf{72.41} \\
\bottomrule
\end{tabular}
}
\vspace{-3mm}
\end{wraptable}

For Sudoku, we compare to the autoregressive model (ARM) of \citet{shah2024causal}, with and without its learned variable-ordering heuristic, and to masked diffusion (MDM) with the MDLM objective~\citep{sahoo2024simple}, reporting vanilla parallel sampling and the adaptive top-probability and top-probability-margin rules of \citet{kim2025train}. For Countdown, we include the baseline published  by
\citet{ye2024beyond}: GPT-2 Scratch, Stream-of-Search, LLaMA, VDM, D3PM, and
RDM, together with the same 6M-parameter MDM decoding variants for a
same-order-of-magnitude diffusion comparison. On both tasks we further include CCDD~\citep{zhou2025coevolutionary}, reproduced at a matched 6M-parameter scale under the same protocol with its original backbone replaced by our own for fair comparison, as a hybrid discrete--continuous baseline. Parameter counts in our tables
refer to sampling-time generative parameters, excluding token embeddings (\cref{app:train}).
For LM1B, we compare against an autoregressive Transformer~\citep{vaswani2017attention}, the baseline discrete diffusion models MDM~\citep{sahoo2024simple}, SEDD~\citep{lou2023discrete} and Duo~\citep{sahoo2025diffusionduality}, and the continuous diffusion models Plaid~\citep{gulrajani2023likelihood} and LangFlow~\citep{chen2026langflow}.
Implementation details %
are given in \cref{app:details}. %

\subsection{Token-Dependent Reasoning: Sudoku}
\label{sec:exp_sudoku}

\leavevmode\begin{wraptable}{r}{0.5\textwidth}
\vspace{-4mm}
\centering
\caption{Countdown accuracy (\%$\uparrow$).
Best overall result per subtask in \textbf{bold}; best result at the 6M parameter scale \underline{underlined}.}
\vspace{-2mm}
\label{tab:countdown}
\setlength{\tabcolsep}{4pt}
{\small
\begin{tabular}{lrrr}
\toprule
Method & \#Params & CD4 & CD5 \\
\midrule
\multicolumn{3}{l}{\textit{Autoregressive}} \\
\multirow{3}{*}{GPT-2 Scratch} & 6M   & 31.9 &  4.3 \\
                               & 85M  & 45.8 &  5.1 \\
                               & 303M & 41.3 &  4.5 \\
Stream-of-Search               & 250M & 54.2 &  --  \\
\multirow{2}{*}{LLaMA} & 7B  & 41.1 &  6.7 \\
                                    & 13B & 51.1 &  7.4 \\
\midrule
\multicolumn{3}{l}{\textit{Discrete diffusion}} \\
VDM   & 85M & 73.4 & 16.3 \\
D3PM  & 85M & 83.1 & 27.6 \\
RDM   & 85M & \textbf{87.0} & \textbf{45.8} \\
MDM (vanilla)           & 6M & 6.7 & 0.3 \\
MDM (top-prob.)         & 6M & 47.4 & 21.4 \\
MDM (top-prob.\ margin) & 6M & 50.8 & 21.3 \\
\midrule
\multicolumn{3}{l}{\textit{Hybrid discrete--continuous}} \\
CCDD & 6M & 81.18 & 25.35 \\
\midrule
\name{} (ours) & 6M & \underline{84.41} & \underline{37.52} \\
\bottomrule
\end{tabular}
}
\vspace{-7mm}
\end{wraptable}
\noindent\textbf{Setup.}
Sudoku puzzles are $9{\times}9$ grids ($L=81$, $K=10$), given as partially filled grids. The model must fill all blank cells, and a prediction counts as correct only when all 81 cells match the unique solution. We follow the data and evaluation protocol of \citet{kim2025train}: the standard split of \citet{shah2024causal} contains puzzles solvable by a fixed set of seven logical strategies, and the hard split consists of the remaining puzzles, which require a strategy outside that set (\cref{app:data}).

\noindent\textbf{Results.}
\cref{tab:sudoku} compares \name{} to autoregressive and diffusion baselines, with the autoregressive and masked-diffusion results taken from \citet{kim2025train}. \name{} is competitive on both splits. On the in-distribution Easy set, it surpasses the best masked-diffusion variant and the much larger 42M-parameter autoregressive model, while the matched CCDD implementation attains comparable and marginally higher accuracy (94.65 vs.\ 94.21). On the out-of-distribution Hard split, \name{} instead leads CCDD (72.41 vs.\ 70.73), indicating that its advantage is most pronounced beyond the solution strategies represented in training.

\subsection{Mathematical Reasoning: Countdown}
\label{sec:exp_countdown}

\noindent\textbf{Setup.}
Countdown~\citep{ye2024beyond} is a mathematical reasoning task that generalizes the Game of 24: given $n$ numbers and a target integer, the model must produce a chain of arithmetic steps that reaches the target exactly. We follow \citet{ye2024beyond}, with 500k problems and 10\% of the target values held out for out-of-distribution evaluation; CD4 and CD5 use four and five input numbers. Since a Countdown puzzle admits many valid solutions, we report answer-correct accuracy (\cref{app:data}).

\noindent\textbf{Results.} 
\cref{tab:countdown} compares \name{} to autoregressive and diffusion baselines. Results for the non-MDM baselines are taken from \citet{ye2024beyond}, while the MDM and CCDD rows are our reproduced 6M-parameter baseline runs under the same Countdown setting. Most diffusion variants outperform substantially larger autoregressive models; following \citet{ye2024beyond}, we read this gap as any-order decoding suiting subgoal-imbalanced planning, not as a property of any single method. The informative comparison for \name{} is therefore with diffusion models of the same order of magnitude, and there it holds a clear advantage: it surpasses the matched hybrid CCDD on both subtasks, with the margin widening on the longer-horizon CD5 (37.52 vs.\ 25.35). %
Much larger discrete diffusion models still achieve the best overall numbers, but \name{} remains highly competitive at a fraction of the parameter cost.

\vspace{-2mm}
\subsection{Language Modeling: LM1B}
\label{sec:exp_lm1b}

\noindent\textbf{Setup.}
To evaluate our method on natural language modeling, we test unconditional generation on the One Billion Word benchmark (LM1B; \citealp{chelba2013one}) at sequence length 128 with 118M parameters in the denoiser and token decoder, excluding embeddings (\cref{app:train}), and report generative perplexity (Gen.\ PPL) under GPT-2-Large~\citep{radford2019language} over 1{,}024 samples at 128 sampling steps, following the protocol of LangFlow~\citep{chen2026langflow}; sampling uses classifier-free guidance with $w=2.75$ (\cref{app:lm1b_cfg}).

\begin{wraptable}{r}{0.5\textwidth}
\vspace{-1mm}
\centering
\caption{Unconditional generation results on LM1B. Baseline results are from models retrained by the LangFlow authors~\citep{chen2026langflow}.}
\vspace{-2mm}
\label{tab:lm1b}
\setlength{\tabcolsep}{4pt}
{\small
\begin{tabular}{lrr}
\toprule
Method & \#Params & Gen. PPL $\downarrow$ \\
\midrule
\multicolumn{3}{l}{\textit{Autoregressive}} \\
Transformer & 108M & 66.7 \\
\midrule
\multicolumn{3}{l}{\textit{Diffusion}} \\
MDM      & 116M & 103.9 \\
SEDD     & 116M & 115.9 \\
Duo      & 116M & 97.6 \\
Plaid    & 109M & 77.3 \\
LangFlow & 117M & 92.2 \\
\midrule
\name{} (ours) & 118M & 75.5 \\
\midrule
\textit{Ground truth} & -- & 40.4 \\
\bottomrule
\end{tabular}
}
\vspace{-4mm}
\end{wraptable}

\noindent\textbf{Results.}
\Cref{tab:lm1b} compares \name{} with autoregressive, discrete-diffusion and continuous-diffusion baselines on LM1B, with baseline results taken from LangFlow~\citep{chen2026langflow}. \name{} achieves the best generative perplexity among the evaluated diffusion models, outperforming all discrete diffusion baselines and both continuous models that decode only once at the end. Entropy remains stable under moderate reductions in NFE and across the CFG sweep (\cref{tab:lm1b_sample_quality,fig:lm1b_cfg}). At large batch sizes, \name{} also maintains lower wall-clock sampling time, suggesting favorable efficiency for batched parallel generation (\cref{fig:lm1b_speed}). Together with the structured-task ablations below, these comparisons are consistent with a benefit from iterative token feedback over decoding the continuous representation only once at the end.

\subsection{Ablation Studies}
\label{sec:ablation}

\leavevmode\begin{wraptable}{r}{0.54\textwidth}
\vspace{-4mm}
\centering
\caption{Component ablation on Sudoku: accuracy (\%$\uparrow$).
\textbf{Cont.\ latent}: latent trajectory $x_t$ present. %
\textbf{Disc.\ cond.}: $k_t$ passed to denoiser $d_\phi$.}
\label{tab:ablation_component}
\setlength{\tabcolsep}{3pt}
{\small
\begin{tabular}{lcccc}
\toprule
Method & \makecell{Cont.\\ latent} & \makecell{Disc.\\ cond.} & Easy & Hard  \\
\midrule
MDM (top-prob.\ margin)      & $\times$     & $\times$     &89.49& 49.88              \\
Latent DM (w/o disc.\ cond.) & $\checkmark$ & $\times$     &50.46& 24.74        \\
\name{} (ours)                  & $\checkmark$ & $\checkmark$ &\textbf{94.21} & \textbf{72.41} \\
\bottomrule
\end{tabular}
}
\vspace{-3mm}
\end{wraptable}
We test %
\name{}'s key design choices and analyze its generation dynamics; unless otherwise specified, ablations use Sudoku.

\noindent\textbf{Framework Choices.}
Our central hypothesis is that the gains come from the hierarchical coupling itself, i.e., a %
latent plan with per-step token readout and scaffold feedback. To test this, we remove each level in turn. %
Latent DM removes the token level %
by dropping the token state $k_t$ from the denoiser $d_\phi$, which leaves a standard continuous latent diffusion model that decodes tokens only at the final step. The MDM baseline from \cref{tab:sudoku} removes the latent level altogether, leaving a purely discrete chain.

As shown in \cref{tab:ablation_component}, removing either level %
degrades performance, particularly on the Hard split where global constraint satisfaction is critical. Latent DM falls far below \name{}, demonstrating that a continuous latent without a discrete scaffold can over-smooth precise token-level constraints. Conversely, MDM's bottleneck highlights the limitations of independent discrete sampling. %

\begin{wraptable}[12]{r}{0.42\textwidth}
\vspace{-4mm}
\centering
\caption{Ablation on Sudoku: discrete noise type and token-ordering heuristic.
Accuracy (\%$\uparrow$). Best result in \textbf{bold}.}
\label{tab:ablation_ordering_noise}
\setlength{\tabcolsep}{6pt}
\resizebox{1\linewidth}{!}{  
\begin{tabular}{llcc}
\toprule
Noise type & Token ordering & Easy & Hard \\
\midrule
\multirow{3}{*}{Absorbing-state}
  & Random              &69.60& 43.74 \\
  & Top-prob.           &74.72& 47.50 \\
  & Top-prob.\ margin   &75.59& 48.25 \\
\midrule
Uniform-state
  & Parallel update     &\textbf{94.21} & \textbf{72.41} \\
\bottomrule
\end{tabular}
}
\vspace{-3mm}
\end{wraptable}
\noindent\textbf{Discrete Forward Process Design.}
We next evaluate the discrete forward process choice: \cref{tab:ablation_ordering_noise} compares the \textit{absorbing} and \textit{uniform} kernels.
For the absorbing kernel, the presence of distinct masked tokens allows for explicit decoding schedules. We compare \textit{random} selection to adaptive strategies: \textit{top-prob.} (unmasking the token with the highest predicted probability) and \textit{top-prob. margin} (top-two probability gap~\citep{kim2025train}). Because the uniform kernel corrupts all tokens uniformly without a separate masked set, we apply the standard parallel update, re-noising all positions through the forward kernel at each step.

\newcommand{\trajectoryfigure}{%
\begin{wrapfigure}[]{r}{0.42\textwidth}\vspace{-8mm}
\centering
\begin{subfigure}[b]{\repwidth\linewidth}
        \centering
        \includegraphics[width=\linewidth]{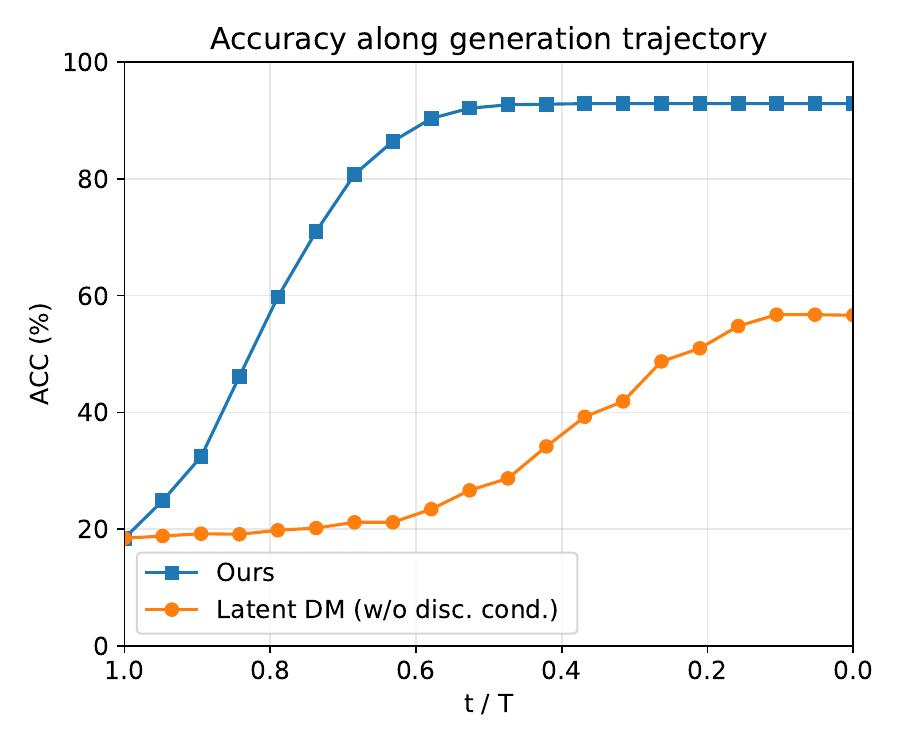}
        \vspace{-6mm}
        \caption{Trajectory decoding.}
        \label{fig:repr_trajectory_sub1}
    \end{subfigure}\subfigureseparator
    \vspace{-2mm}
    \begin{subfigure}[b]{\repwidth\linewidth}
        \centering
        \includegraphics[width=\linewidth]{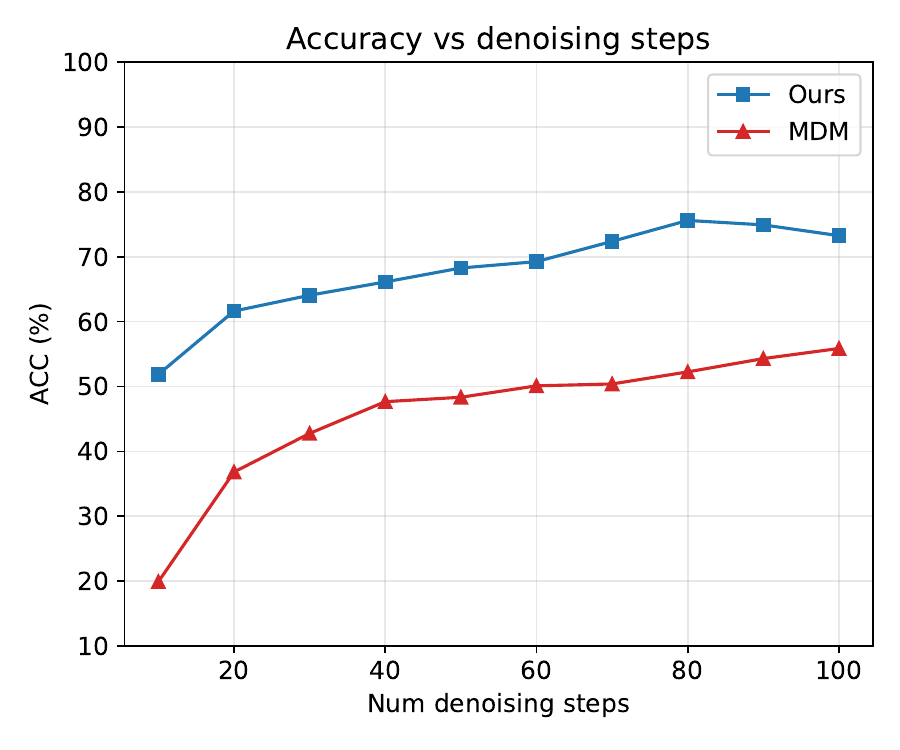}
        \vspace{-6mm}
        \caption{Step-count robustness.}
        \label{fig:repr_trajectory_sub2}
    \end{subfigure}
    \vspace{-4mm}
\caption{%
\textbf{(a)} Accuracy of the decoded intermediate prediction $\hat{x}_0$ at each normalized step $t/T$; \name{} improves faster. \textbf{(b)} Final accuracy on Hard Sudoku vs.\ step count; \name{} stays robust with fewer steps. %
}
\label{fig:repr_trajectory}
\vspace{-5mm}\end{wrapfigure}
}
\trajectoryfigure
The results demonstrate that while adaptive ordering modestly improves the absorbing kernel, the uniform kernel is substantially stronger for our hierarchical sampler. We therefore use the uniform kernel as the default in all main experiments. Two properties of the hierarchy explain this gap: the discrete state serves the latent denoiser as an informative conditioning scaffold, which absorbing corruption empties of content, and per-step readout keeps every position revisable, which adaptive unmasking forecloses; \cref{app:uniform} %
expands this analysis.

\noindent\textbf{Information Emergence.} To understand why \name{} outperforms the baselines, we analyze how information crystallizes along the reverse trajectory: at each step $t$ we decode the intermediate clean-latent prediction $\hat{x}_0$ into tokens and measure the resulting solution accuracy. As shown in \cref{fig:repr_trajectory_sub1}, accuracy is near chance at $t/T = 1.0$, where the latent is near Gaussian noise, but rises significantly earlier and reaches a much higher plateau than Latent DM. Conditioning the continuous denoiser on the %
discrete scaffold $k_t$ ``locks in'' structural constraints  sooner, refining the solution progressively through the feedback loop rather than resolving it at the final boundary step.

\noindent\textbf{Inference Efficiency.} We further evaluate accuracy across varying numbers of denoising steps on the Hard Sudoku split. \cref{fig:repr_trajectory_sub2} illustrates that \name{} is  robust to reduced step counts, maintaining strong predictive accuracy. %
In contrast, the purely discrete MDM baseline degrades more sharply as the number of steps decreases. This pattern is consistent with parallel MDM predictions relying on repeated refinement to compensate for their conditional-independence approximation. In \name{}, all parallel readouts are tied to one shared latent state, and the scaffold conditions the latent update inside every step. \Cref{app:inference_time} %
further shows that our implementation is faster in wall-clock time than the reproduced MDM implementation at a comparable parameter scale.%

\section{Related Work}
\label{sec:related}

\noindent\textbf{Discrete diffusion language models.}
D3PM~\citep{austin2021structured}, SEDD~\citep{lou2023discrete}, MDLM~\citep{sahoo2024simple}, RDM~\citep{zheng2023reparameterized} and LLaDA~\citep{nie2025large} introduced absorbing- and uniform-state diffusion over tokens from small to 8B scale, with block diffusion~\citep{arriola2025block} and discrete flow matching~\citep{gat2024discrete} as alternative decompositions. MGDM~\citep{ye2024beyond} and \citet{kim2025train} established their advantage on planning tasks and the role of decoding order. \name{} preserves parallel token updates and adds a continuous variable. %

\noindent\textbf{Continuous diffusion for text.}
Continuous models denoise a continuous state of the sequence, either per-token embeddings~\citep{li2022diffusion, gong2022diffuseq, gulrajani2023likelihood, chen2026langflow, hu2026elf, shen2026codar} or an encoder-compressed latent~\citep{zhang2023planner, lovelace2023latent, guo2026continuous, meshchaninov2026train}, with a denoiser that takes only the continuous state as input and tokens decoded from the clean result. \name{} reads tokens out of the latent at every step and feeds them back into its denoiser, so the two levels refine each other throughout generation (\cref{sec:model}).

\noindent\textbf{Hybrid discrete--continuous diffusion.}
Hybrid models pair a discrete chain with a continuous channel, whether a global latent drawn once (VMD)~\citep{zhang2025variational}, per-token hints (CADD)~\citep{zheng2025continuously}, or a parallel embedding chain (CCDD)~\citep{zhou2025coevolutionary}, and in these designs the tokens keep a transition chain of their own. Further couplings are explored by \citet{shariatian2026latent}, \citet{pynadath2025candi} and \citet{lemercier2026diladiff}. 
\name{} instead reads tokens out of it at every step, so the tokens carry no transition kernel of their own. 
See \cref{app:comparison} for more. %

\section{Conclusion}
\label{sec:conclusion}

We proposed \name{}, a diffusion framework for discrete sequences whose reverse process is a two-level chain, with tokens read out of a continuous latent at every step and fed back into its denoiser.
A single variational lower bound yields the coupled objective, and the hierarchical coupling %
addresses the token-independence bottleneck of parallel decoding by routing cross-token dependence through a compact learned latent. %
Experiments on Sudoku, Countdown, and LM1B show that the framework applies across globally constrained generation and unconditional language modeling. Structured-task comparisons show that scaffold feedback is important and that the coupled model outperforms continuous-only and  discrete alternatives. %

\subsubsection*{Acknowledgments}
Work supported in part by NSF grants 2008387, 2045586,
2106825, 2551476, MRI 1725729, NIFA award 2020-67021-32799 and Amazon AI PhD Fellowship.

\bibliography{refs}
\bibliographystyle{conference_style}

\clearpage
\appendix
\section*{Appendix}
\crefalias{section}{appendix}
\crefalias{subsection}{appendix}
\crefalias{subsubsection}{appendix}

\section{Proof of \cref{prop:elbo}}
\label{app:proof}

We derive the ELBO step by step, starting from the definitions in \cref{sec:model}.

\paragraph{Step 1: Jensen's inequality.}
Starting from the marginal likelihood of $k_0$, we make use of the variational distribution given in \cref{eq:forward} and apply Jensen's inequality:
\begin{align}
\log p(k_0)
&= \log \int \sum_{k_{1:T}}
   p_{\theta,\phi}(k_{0:T}, x_{0:T})\,dx_{0:T}
\notag\\
&= \log \E_{q_\psi(x_{0:T}, k_{1:T} \mid k_0)}\!\left[
  \frac{p_{\theta,\phi}(k_{0:T}, x_{0:T})}
       {q_\psi(x_{0:T}, k_{1:T} \mid k_0)}
\right]
\notag\\
&\geq
\E_{q_\psi(x_{0:T}, k_{1:T} \mid k_0)}\!\left[
  \log\frac{p_{\theta,\phi}(k_{0:T}, x_{0:T})}
           {q_\psi(x_{0:T}, k_{1:T} \mid k_0)}
\right]
=: \calL(k_0).
\end{align}
For readability, we use $\E_{q_\psi}$ to denote the expectation under $q_\psi(x_{0:T}, k_{1:T}\mid k_0)$ in what follows.

\paragraph{Step 2: Substituting the factorizations.}
Substituting \cref{eq:gen_model} and \cref{eq:forward} and taking logs:
\begin{align}
\calL(k_0)
&=
\E_{q_\psi}\Bigl[
  \log p(x_T)
  + \log p_\theta(k_0 \mid x_0)
  + \textstyle\sum_{t=1}^T \log p_\theta(k_t \mid x_t)
  + \textstyle\sum_{t=1}^T \log p_\phi(x_{t-1} \mid x_t, k_t)
\notag\\
&\qquad\quad
  - \log q_\psi(x_0 \mid k_0)
  - \textstyle\sum_{t=1}^T \log q(x_t \mid x_{t-1})
  - \textstyle\sum_{t=1}^T \log q(k_t \mid k_{t-1})
\Bigr].
\label{eq:elbo_expanded}
\end{align}
We separate the $t=1$ boundary term from the $t \ge 2$ terms:
\begin{align}
\calL(k_0)
&=
\E_{q_\psi}\Bigl[
  \log p_\theta(k_0 \mid x_0)
  - \log q_\psi(x_0 \mid k_0)
\Bigr]
\notag\\
&\quad+
\E_{q_\psi}\Bigl[
  \log p_\theta(k_1 \mid x_1)
  + \log p_\phi(x_0 \mid x_1, k_1)
  - \log q(x_1 \mid x_0)
  - \log q(k_1 \mid k_0)
\Bigr]
\notag\\
&\quad+
\sum_{t=2}^T
\E_{q_\psi}\Bigl[
  \log p_\theta(k_t \mid x_t)
  + \log p_\phi(x_{t-1} \mid x_t, k_t)
  - \log q(x_t \mid x_{t-1})
  - \log q(k_t \mid k_{t-1})
\Bigr]
\notag\\
&\quad+
\E_{q_\psi}\bigl[\log p(x_T)\bigr].
\label{eq:elbo_split}
\end{align}

\paragraph{Step 3: Telescoping on continuous terms.}
For $t \ge 2$, we apply the diffusion posterior identity:
\begin{equation}
q(x_t \mid x_{t-1})
=
q(x_{t-1} \mid x_t, x_0)\,
\frac{q(x_t \mid x_0)}{q(x_{t-1} \mid x_0)},
\end{equation}
which gives
\begin{equation}
\log q(x_t \mid x_{t-1})
=
\log q(x_{t-1} \mid x_t, x_0)
+ \log q(x_t \mid x_0)
- \log q(x_{t-1} \mid x_0).
\end{equation}
Substituting this into the sum over $t=2,\dots,T$ in \cref{eq:elbo_split}, the terms $-\log q(x_t \mid x_0) + \log q(x_{t-1} \mid x_0)$ telescope:
\begin{equation}
\sum_{t=2}^T
\bigl(
-\log q(x_t \mid x_0)
+
\log q(x_{t-1} \mid x_0)
\bigr)
=
-\log q(x_T \mid x_0)
+
\log q(x_1 \mid x_0).
\end{equation}
The resulting $+\log q(x_1 \mid x_0)$ cancels the $-\log q(x_1 \mid x_0)$ in the $t{=}1$ boundary term. After telescoping:
\begin{align}
\calL(k_0)
&=
\E_{q_\psi}\Bigl[
  \log p_\theta(k_0 \mid x_0)
  - \log q_\psi(x_0 \mid k_0)
\Bigr]
\notag\\
&\quad+
\E_{q_\psi}\Bigl[
  \log p_\theta(k_1 \mid x_1)
  + \log p_\phi(x_0 \mid x_1, k_1)
  - \log q(k_1 \mid k_0)
\Bigr]
\notag\\
&\quad+
\sum_{t=2}^T
\E_{q_\psi}\bigl[
  \log p_\theta(k_t \mid x_t)
  + \log p_\phi(x_{t-1} \mid x_t, k_t)
\notag\\
&\hspace{5em}
  - \log q(x_{t-1} \mid x_t, x_0)
  - \log q(k_t \mid k_{t-1})
\bigr]
\notag\\
&\quad+
\E_{q_\psi}\bigl[
  \log p(x_T) - \log q(x_T \mid x_0)
\bigr].
\label{eq:elbo_after_telescoping}
\end{align}
The last line is the negative prior KL:
\begin{equation}
\E_{q_\psi}\bigl[\log p(x_T) - \log q(x_T \mid x_0)\bigr]
=
-\E_{q_\psi(x_0 \mid k_0)}\!\Bigl[\KL\bigl(q(x_T \mid x_0)\,\|\,p(x_T)\bigr)\Bigr].
\end{equation}

\paragraph{Step 4: Rewriting as KL divergences.}

\emph{Boundary term ($t=1$).}
Rearranging the $t{=}1$ terms in \cref{eq:elbo_after_telescoping}:
\begin{align}
&\E_{q_\psi}\Bigl[
  \log p_\theta(k_1 \mid x_1)
  + \log p_\phi(x_0 \mid x_1, k_1)
  - \log q(k_1 \mid k_0)
\Bigr]
\notag\\
&=
\E_{q_\psi(x_0, x_1 \mid k_0),\,q(k_1 \mid k_0)}
\bigl[\log p_\phi(x_0 \mid x_1, k_1)\bigr]
-
\E_{q_\psi(x_1 \mid k_0)}
\!\left[
  \KL\!\bigl(q(k_1 \mid k_0)\,\|\,p_\theta(k_1 \mid x_1)\bigr)
\right],
\end{align}
where we used the definition of the KL divergence for the discrete part. The remaining term from \cref{eq:elbo_after_telescoping},
$-\E_{q_\psi(x_0\mid k_0)}[\log q_\psi(x_0\mid k_0)]$, is the encoder entropy $H(q_\psi(x_0\mid k_0))$. Together with the reconstruction term and the prior-matching term, this gives terms (i)--(iii) in \cref{eq:elbo}.

\emph{Intermediate terms ($t \ge 2$).}
For each $t \ge 2$, define
\begin{equation}
\mathcal{A}_t
:=
\E_{q_\psi}\bigl[
  \log p_\theta(k_t \mid x_t)
  + \log p_\phi(x_{t-1} \mid x_t, k_t)
  - \log q(x_{t-1} \mid x_t, x_0)
  - \log q(k_t \mid k_{t-1})
\bigr].
\end{equation}
This can be rewritten as an expectation of a log-ratio over the joint variational distribution $q(x_{t-1}, k_t \mid x_t, x_0, k_{t-1}) = q(x_{t-1} \mid x_t, x_0)\,q(k_t \mid k_{t-1})$:
\begin{align}
\mathcal{A}_t
&=
\E_{q_\psi(x_0, x_t, k_{t-1} \mid k_0)}
\E_{q(x_{t-1} \mid x_t, x_0)\,q(k_t \mid k_{t-1})}\!
\left[
\log
\frac{
p_\phi(x_{t-1} \mid x_t, k_t)\, p_\theta(k_t \mid x_t)
}{
q(x_{t-1} \mid x_t, x_0)\, q(k_t \mid k_{t-1})
}
\right]
\notag\\
&=
-
\E_{q_\psi(x_0, x_t, k_{t-1} \mid k_0)}
\Bigl[
\KL\bigl(
  q(x_{t-1} \mid x_t, x_0)\,q(k_t \mid k_{t-1})
  \|
  p_\phi(x_{t-1} \mid x_t, k_t)\,p_\theta(k_t \mid x_t)
\bigr)
\Bigr].
\label{eq:At_joint_kl}
\end{align}
Combining terms (i)--(iv) gives \cref{prop:elbo}. \qed

\paragraph{Chain rule decomposition (\cref{eq:kl_decompose}).}

We now derive the decomposition of the joint KL in term~(iv) into the two complementary learning signals shown in \cref{eq:kl_decompose}.

Abbreviate $Q_{t-1} := q(x_{t-1} \mid x_t, x_0)$, $Q_t^k := q(k_t \mid k_{t-1})$, $P_{t-1} := p_\phi(x_{t-1} \mid x_t, k_t)$, $P_t^k := p_\theta(k_t \mid x_t)$. The joint KL expands by definition:
\begin{align}
&\KL\!\bigl(Q_t^k \, Q_{t-1} \,\big\|\, P_t^k \, P_{t-1}\bigr)
\notag\\*
&=
\E_{Q_t^k \, Q_{t-1}}\!\left[
  \log Q_t^k + \log Q_{t-1} - \log P_t^k - \log P_{t-1}
\right]
\notag\\*
&=
\E_{Q_t^k}\!\left[\log Q_t^k - \log P_t^k\right]
+
\E_{Q_t^k}\!\left[
  \E_{Q_{t-1}}\!\bigl[\log Q_{t-1} - \log P_{t-1}\bigr]
\right],
\end{align}
where the split uses the fact that $Q_{t-1} = q(x_{t-1}\mid x_t, x_0)$ does not depend on $k_t$, while $P_{t-1} = p_\phi(x_{t-1}\mid x_t, k_t)$ does. Therefore:
\begin{align}
\KL\!\bigl(Q_t^k\,Q_{t-1}\,\big\|\,P_t^k\,P_{t-1}\bigr)
&=
\underbrace{\KL\!\bigl(q(k_t \mid k_{t-1})\,\|\,p_\theta(k_t \mid x_t)\bigr)}_{\text{discrete token prediction}}
\notag\\
&\quad+
\underbrace{
  \E_{q(k_t \mid k_{t-1})}\!\left[
    \KL\!\bigl(q(x_{t-1} \mid x_t, x_0)\,\|\,p_\phi(x_{t-1} \mid x_t, k_t)\bigr)
  \right]
}_{\text{continuous denoising}},
\end{align}
which is exactly \cref{eq:kl_decompose}. The key observation is that the chain rule of the KL divergence,
\begin{equation}
\KL\bigl(p(A)\,p(B\mid A)\|q(A)\,q(B\mid A)\bigr)
= \KL\bigl(p(A)\|q(A)\bigr)
  + \E_{p(A)}\bigl[\KL\bigl(p(B\mid A)\|q(B\mid A)\bigr)\bigr],
\end{equation}
applies here with $A = k_t$ and $B = x_{t-1}$: the ``forward'' distribution of $x_{t-1}$ given $k_t$ is $q(x_{t-1}\mid x_t, x_0)$ (constant in $k_t$), while the generative distribution is $p_\phi(x_{t-1}\mid x_t, k_t)$ (dependent on $k_t$). \qed

\paragraph{Step 5: Practical parameterization.}
The ELBO above is derived for a Markov continuous noising chain with tractable Gaussian posteriors $q(x_{t-1}\mid x_t,x_0)$.
Concretely, write the marginal as
\begin{equation}
    q(x_t\mid x_0)
    =
    \mathcal{N}\!\left(
      \sqrt{\bar{\alpha}^{x}_t}\,x_0,\,
      (1-\bar{\alpha}^{x}_t)I
    \right),
    \quad
    x_t=\sqrt{\bar{\alpha}^{x}_t}\,x_0+\sqrt{1-\bar{\alpha}^{x}_t}\,\epsilon,
    \quad \epsilon\sim\mathcal{N}(0,I),
    \label{eq:gaussian_forward}
\end{equation}
so that $q(x_{t-1}\mid x_t,x_0)=\mathcal{N}(\tilde{\mu}_t(x_t,x_0),\tilde{\Sigma}_t)$.
If $p_\phi(x_{t-1}\mid x_t,k_t)=\mathcal{N}(\mu_\phi(x_t,k_t,t),\sigma_t^2I)$, then
\begin{equation}
    \KL\!\bigl(
      q(x_{t-1}\mid x_t,x_0)
      \,\|\,p_\phi(x_{t-1}\mid x_t,k_t)
    \bigr)
    =
    \frac{1}{2\sigma_t^2}
    \bigl\|\mu_\phi(x_t,k_t,t)-\tilde{\mu}_t(x_t,x_0)\bigr\|^2
    + C_t,
    \label{eq:gaussian_mean_mse}
\end{equation}
where $C_t$ is independent of learned parameters.
With the usual noise-prediction parameterization, this is equivalent to a weighted MSE
\begin{equation}
    \calJ_t^{\mathrm{cont}}
    =
    \E_{q_\psi(x_0,x_t,k_t\mid k_0)}
    \!\left[
      \omega_t
      \bigl\|\epsilon_\phi(x_t,k_t,t/T)-\epsilon\bigr\|^2
    \right],
    \label{eq:gaussian_noise_mse}
\end{equation}
for a known schedule-dependent weight $\omega_t$~\citep{ho2020denoising,luo2022understand}.
For the discrete token prediction term, the categorical kernel gives $\KL(q(k_t \mid k_{t-1})\|p_\theta(k_t \mid x_t)) = \E_{q(k_t|k_{t-1})}[-\log p_\theta(k_t \mid x_t)] + C'$ with constant $C'$ independent of $\theta$.
After averaging over $q(k_{t-1}\mid k_0)$, the trainable cross-entropy part is equivalently
\begin{equation}
    \E_{q(k_t\mid k_0)}
    \bigl[-\log p_\theta(k_t\mid x_t)\bigr],
    \label{eq:noisy_token_ce}
\end{equation}
again up to constants independent of $\theta$.

\paragraph{Clean-token surrogate for the discrete KL.}
To optimize the discrete part of the ELBO we must evaluate the noisy-token cross-entropy in \cref{eq:noisy_token_ce}, which requires the noisy-token distribution $p_\theta(k_t\mid x_t)$ at every noise level $t$. The challenge is that this is one head per noise level, while the rest of the objective only ever supervises the model at the clean endpoint. We highlight two obstacles, then resolve both with a single $x_0$-prediction parameterization that ties everything back to the clean-token predictor $p_\theta(k_0\mid \cdot)$ already used in term~(i).

\emph{Obstacle 1: a per-$t$ predictor is wasteful and inconsistent.} The underlying signal at every $t$ is the same clean sequence $k_0$ passed through the known categorical kernel $q(k_t\mid k_0)$ from \cref{eq:discrete_forward}, so a separate head per $t$ would relearn information that is already analytically available. It is also inconsistent with the clean-endpoint head $p_\theta(k_0\mid x_0)$ supervised by term~(i): we would maintain two discrete heads for what is essentially the same task.

We therefore reuse a single, time-invariant clean-token predictor $p_\theta(k_0\mid \cdot)$, the same head appearing in $\calJ_{\mathrm{recon}}$, and compose it with the known forward kernel from \cref{eq:discrete_forward}:
\begin{equation}
    p_\theta(k_t\mid x_t)
    =
    \sum_{\tilde{k}_0}
    q(k_t\mid \tilde{k}_0)\,
    p_\theta(\tilde{k}_0\mid \hat{x}_0(x_t)),
    \label{eq:appendix_x0_pred_marginal}
\end{equation}
where $\hat{x}_0(x_t)$ is the clean-latent estimate produced by the denoiser. Its explicit form depends on the parameterization of $d_\phi$. In our implementation the denoiser predicts the clean latent directly, $\hat{x}_0(x_t) = d_\phi(x_t,k_t,t/T)$, an $x$-prediction parameterization (see \cref{app:arch}); an equivalent velocity parameterization would instead set $\hat{x}_0(x_t)=x_t-(t/T)\,v_\phi(x_t,k_t,t/T)$. The data-processing argument that follows does not rely on either form. With this parameterization the noisy-token distribution at every $t$ is determined in closed form once $p_\theta(k_0\mid\cdot)$ is learned, and a single head is shared across $t=0,1,\dots,T$.

\emph{Obstacle 2: directly optimizing the noisy-token KL gives a smeared gradient.} Even after \cref{eq:appendix_x0_pred_marginal}, optimizing $-\log p_\theta(k_t\mid x_t)$ requires differentiating through the mixture induced by $q(k_t\mid k_0)$. Under the absorbing kernel, in particular, $q(k_t\mid k_0)$ maps many distinct candidate clean tokens to the same noisy symbol \textsc{[mask]}, so the gradient is averaged over a large equivalence class and the per-position learning signal on $p_\theta(k_0\mid\cdot)$ is diffuse. To recover a sharp per-position cross-entropy, and to match the functional form of term~(i), we replace the noisy-token KL by a clean-token upper bound obtained from the data-processing inequality.

For a fixed clean sequence $k_0$, the noisy-token KL induced by the model family in \cref{eq:appendix_x0_pred_marginal} is
\begin{equation}
    \KL\!\left(
      q(k_t\mid k_0)
      \,\middle\|\,
      \textstyle\sum_{\tilde{k}_0} q(k_t\mid \tilde{k}_0)\,p_\theta(\tilde{k}_0\mid \hat{x}_0(x_t))
    \right).
    \label{eq:noisy_kl_x0_pred}
\end{equation}
Both arguments are obtained by pushing a clean-token distribution ($\delta_{k_0}$ on the left, $p_\theta(\cdot\mid\hat{x}_0(x_t))$ on the right) through the same channel $q(k_t\mid k_0)$, so the data-processing inequality gives
\begin{equation}
    \eqref{eq:noisy_kl_x0_pred}
    \le
    \KL\!\left(
      \delta_{k_0}
      \,\middle\|\,
      p_\theta(\cdot\mid \hat{x}_0(x_t))
    \right)
    =
    -\log p_\theta(k_0\mid \hat{x}_0(x_t)).
    \label{eq:clean_ce_upper_bound}
\end{equation}
Thus the clean-token cross-entropy upper-bounds the trainable noisy-token KL, up to entropy terms determined only by the forward kernel.
For the absorbing kernel, if the token predictor places probability mass only on non-mask data tokens, this relation tightens to the usual masked-diffusion form: the trainable part of \cref{eq:noisy_token_ce} is $\alpha_t\bigl[-\log p_\theta(k_0\mid \hat{x}_0(x_t))\bigr]$, with all remaining terms independent of $\theta$.

In the practical objective we further supervise the same token predictor at the clean endpoint, replacing $\hat{x}_0(x_t)$ by the variational clean latent $x_0$ to obtain
\begin{equation}
    \calJ_{\mathrm{recon}}
    =
    \E_{q_\psi(x_0\mid k_0)}
    \bigl[-\log p_\theta(k_0\mid x_0)\bigr],
    \label{eq:appendix_recon_surrogate}
\end{equation}
while the continuous denoising loss separately drives $\hat{x}_0(x_t)$ toward $x_0$ along the reverse trajectory.
This replacement is a practical surrogate for the discrete KL terms rather than an additional exact algebraic identity.
These substitutions yield the compact practical objective in \cref{eq:obj_simplified}.
In our implementation, we then replace the Gaussian noise-prediction term with a token-conditioned flow-matching interpolant in $x$-prediction form.
Using $s=t/T$, the linear interpolant is
\begin{equation}
    x_t = (1-s)\,x_0 + s\,\epsilon,
    \quad \epsilon\sim\mathcal{N}(0,I),
    \label{eq:fm_forward}
\end{equation}
and the implemented regression loss targets the clean latent,
\begin{equation}
    \calJ_{\mathrm{cont}}
    =
    \sum_{t=1}^{T}
    \E_{q_\psi(x_0, x_t, k_t \mid k_0)}
    \left[
      \|d_\phi(x_t, k_t, t/T) - x_0\|^2
    \right].
    \label{eq:cfm_hdlm}
\end{equation}
Under \cref{eq:fm_forward}, $x_0$ and $\epsilon-x_0$ are related by an affine reparameterization of $(x_t, t)$, so regressing the clean endpoint is equivalent to regressing the CFM velocity target $\epsilon - x_0$ up to a schedule-dependent reweighting. This is a practical surrogate for the continuous denoising term, not an additional exact algebraic step in the ELBO~\citep{lipman2022flow}.
This final replacement gives the implemented $\calJ_{\mathrm{cont}}$ used in \cref{eq:obj_simplified}. \qed

\section{Experimental Details}
\label{app:details}

\subsection{Architecture}
\label{app:arch}

\paragraph{Overview.}
The practical objective in \cref{eq:obj_simplified} is implemented with three jointly trained transformer modules: an encoder, a token predictor, and a continuous denoiser.
The reconstruction and entropy terms require a variational encoder whose latent remains decodable, while the continuous denoising term requires a denoiser network that conditions on both the noisy latent and the current token state.
The interaction between these modules during training is shown in \cref{fig:train}.

\paragraph{Encoder.}
The encoder $q_\psi(x_0\mid k_0)$ is a bidirectional transformer that maps the token sequence $k_0$ to a continuous sequence $x_0\in\mathbb{R}^{M\times d}$.
We use a fully factorized Gaussian parameterization
$q_\psi(x_0\mid k_0)=\mathcal{N}\!\bigl(\mu_\psi(k_0),\,\mathrm{diag}(\sigma_\psi^2(k_0))\bigr)$.
The final hidden states are projected to per-position mean $\mu_\psi(k_0)$ and log-variance $\log\sigma_\psi^2(k_0)$ via two independent linear heads; continuous latents are sampled by the reparameterization trick applied position-wise.
This Gaussian form gives a closed-form $\log q_\psi$ for the entropy regularizer $\mathcal{J}_{\mathrm{ent}}$.

\paragraph{Denoiser.}
The continuous denoiser $d_\phi(x_t,k_t,t/T)$ uses an $x$-prediction parameterization in our flow-matching implementation: it outputs the clean-latent estimate $\hat{x}_0(x_t)$ directly and is regressed against $x_0$ as in \cref{eq:cfm_hdlm}, which under the linear interpolant is equivalent to velocity regression up to a schedule-dependent reweighting.
Its inputs are formed by concatenating the continuous state $x_t\in\mathbb{R}^{M\times d}$ and the learned token embeddings of the discrete state $k_t\in\mathbb{R}^{L\times d}$ along the sequence dimension, yielding a joint sequence of length $M+L$ that is processed by the transformer's self-attention.
The normalized timestep $t/T$ is injected via adaptive layer norm (adaLN) following~\citet{peebles2023scalable}, with scale and shift parameters produced by a two-layer MLP applied to a sinusoidal embedding of $t/T$.

\paragraph{Token predictor.}
Rather than reading the noisy state directly, the token predictor is implemented as a lightweight transformer that consumes a clean latent and emits per-position logits over $\mathcal{V}$, which defines $p_\theta(k_0\mid\cdot)$. At inference it reads the denoiser estimate
\[
    \hat{x}_0(x_t):=d_\phi(x_t,k_t,t/T),
\]
and during training it reads the encoder sample $x_0$, following $\calJ_{\mathrm{recon}}$ in \cref{eq:obj_simplified}. The cross-entropy therefore trains only the encoder and the predictor, and carries no gradient into the denoiser. %
The noisy-token distribution is obtained by marginalizing this clean-token distribution through the discrete forward kernel from \cref{eq:discrete_forward}, i.e., 
\begin{equation}
    p_\theta(k_t \mid x_t) \;=\; \sum_{k_0} q(k_t \mid k_0)\,p_\theta\!\bigl(k_0 \mid \hat{x}_0(x_t)\bigr).
    \label{eq:x_pred_marginal}
\end{equation}
This is an exact marginal for the model family that first decodes a clean sequence and then applies the known categorical noising kernel.
It mirrors the conditional independence structure of \cref{eq:forward}: given $k_0$, the discrete corruption does not require additional information from $x_t$.
Because the token predictor always sees an estimate of the clean latent, it is not conditioned on $t$ and shares no parameters with the encoder or denoiser.

\subsection{Conditional Generation}
\label{app:cond_gen}

For a condition $c$ (e.g., puzzle clues, a class label, or a target sequence property), the conditional model follows by replacing each \emph{learned} component with its conditioned counterpart while leaving the unconditional forward kernels untouched:
\begin{align}
    p_{\theta,\phi}(k_{0:T},\,x_{0:T}\mid c)
    &=
    p(x_T)\;p_\theta(k_0\mid x_0,c)
    \prod_{t=1}^{T}
    p_\theta(k_t\mid x_t,c)\;
    p_\phi(x_{t-1}\mid x_t, k_t, c),
    \label{eq:gen_model_cond}\\
    q_\psi(x_{0:T},\,k_{1:T}\mid k_0, c)
    &=
    q_\psi(x_0\mid k_0,c)
    \prod_{t=1}^{T}
    q(x_t\mid x_{t-1})\;
    q(k_t\mid k_{t-1}).
    \label{eq:forward_cond}
\end{align}
Substituting \cref{eq:gen_model_cond,eq:forward_cond} into the derivation of \cref{prop:elbo} yields the conditional bound $\log p(k_0\mid c)\geq\mathcal{L}(k_0,c)$ with the same four-term structure as \cref{eq:elbo}, with every learned distribution conditioned on $c$.

The practical objective is obtained by the same substitutions in \cref{eq:obj_simplified}: the reconstruction term uses $p_\theta(k_0\mid x_0,c)$, the continuous denoising term uses $d_\phi(x_t,k_t,t/T,c)$ in our flow-matching implementation, and the entropy term uses $q_\psi(x_0\mid k_0,c)$.
At inference time, we use the alternating sampler from \cref{sec:inference} with these conditioned networks.
In implementation, $c$ is injected into all three learned components (the encoder $q_\psi$, the token predictor $p_\theta$, and the continuous denoiser $d_\phi$) as an additional input embedding fed alongside their existing inputs.
We keep the three-argument signature $d_\phi(x_t,k_t,t/T)$ in the main text for notational consistency, with $c$ understood to be appended to the discrete-state stream when conditioning is present.
This is exactly the conditioning prescribed by the conditional ELBO and requires no architectural changes beyond the extra embedding.

\subsection{Data and Serialization Details}
\label{app:data}

\paragraph{Sudoku.}
Puzzles are drawn from the dataset of \citet{radcliffe2020sudoku} and split as in \citet{kim2025train}. The standard split follows \citet{shah2024causal}, who select puzzles solvable by a fixed set of seven logical strategies without backtracking (1.8M training, 0.1M test puzzles); the hard split consists of the complementary 1.1M puzzles, which require a strategy outside this set, and we evaluate on 0.1M of them. Per-puzzle clue counts are nearly identical across the two splits (mean 24.2 vs.\ 24.5, range 20--31), so difficulty arises from solution structure rather than from the number of givens, and each puzzle has a unique solution. We report exact-solution accuracy (\%). Quizzes and solutions are serialized as 81-token sequences over $\{1,\ldots,9\}$, with 0 marking blank cells.

\paragraph{Countdown.}
Problems are generated following \citet{ye2024beyond} with targets from 10 to 100. CD4 and CD5 use four and five input numbers; CD5 requires a longer calculation chain and has a much larger search space. A prediction counts as correct only if the entire decoded chain is valid: it uses only the four permitted arithmetic operations, does not use any provided number more than once, uses only available intermediate results, permits division only when it is exact, and ends with the target value. Agreement with the dataset solution is not required. Each example is serialized as a prompt ``$n_1,\ldots,n_k$ = target'' (e.g., ``$58,84,48,62=96$'') and a response that is a comma-separated chain of sub-equations (e.g., ``$62-58=4,48/4=12,84+12=96$''). Tokens are drawn from a 17-symbol vocabulary (digits 0--9, operators $+,-,\times,\div,=$, comma, and PAD).

\subsection{Training Details}
\label{app:train}
\paragraph{Sudoku, Countdown.}
Our latent flow‑matching model has three modules, all operating at hidden size 256. We use a single latent token ($M=1$) in the Sudoku and Countdown experiments; an ablation over $M$ is provided in \cref{app:m_ablation}. The encoder ($\approx$ 8M parameters) consists of 8 DiT blocks following \citet{arriola2025block}, in which learnable latent queries cross‑attend to the embedded input sequence and produce latent tokens of dimension 256. The continuous denoiser ($\approx$ 4.9M) uses the DiT backbone from \citet{sahoo2025diffusionduality} with 8 blocks, applying concatenated self‑attention over $[c, x_t, k_t]$. We train with continuous‑time flow matching under a linear schedule and use 50 Euler steps for the main structured-task results. The token predictor ($\approx$ 1.7M) consists of 2 DiT blocks following \citet{arriola2025block}, with concatenated self‑attention over $[c, \hat{x}_0]$. Both the token predictor and the denoiser share AdaLN and attention parameters across blocks following DiT-Air~\citep{chen2025ditair}. The sampling-time generative parameters reported in the main tables comprise the denoiser and the token predictor; the encoder is used only during training. For our implementation of CCDD, we adopt its pipeline while replacing its original backbone with our continuous denoiser for fair comparison. Following CCDD, the token sequence is noised at the same level t as the latent, embedded by a separate table, and concatenated into the denoiser's self-attention stream together with the condition and the noisy latent; a second adaLN-zero head reads the token positions of the same residual stream and predicts the clean tokens, so one forward pass emits the continuous and discrete predictions in parallel. Its cross-entropy is added to our objective with a loss weight of 0.2, with all other losses and hyperparameters unchanged. 

We use Adam with learning rate $1{\times}10^{-4}$, batch size 256 per GPU over 7 NVIDIA L40s GPUs. We set $\lambda_\text{recon}=1.0, \lambda_\text{ent}=8{\times}10^{-3}, \lambda_\text{cont}=0.5$, dropout rate $0.1$. Additionally, an Exponential Moving Average (EMA) was applied to the model weights with a decay factor of $0.999$. For all tasks, we additionally use a BERT-style reconstruction loss with weight 0.2. Each input token is independently masked with probability 0.15; the encoder receives the resulting partially masked sequence, and the token decoder is trained to recover the original clean tokens. We repeat each of our reported configurations over multiple independent runs to verify stability.

\paragraph{LM1B.}
The LM1B model uses the same encoder--denoiser--predictor design and sampling procedure as the structured-task models, but scales the flow backbone to a 107.8M-parameter DiT-small with hidden size 768, 14 blocks, and 12 attention heads. The 118M parameters reported in \cref{tab:lm1b} comprise the sampling-time denoiser and token decoder, excluding token embeddings and the training-only encoder; the complete training model has approximately 200.5M parameters. We use the Qwen3 tokenizer~\citep{yang2025qwen3} ($|\mathcal{V}|=151{,}669$) at sequence length 128. The LangFlow baselines use the bert-base-uncased tokenizer at the same length; GPT-2-Large scores the detokenized text, placing all models in a common evaluation space. At inference, we generate 1{,}024 unconditional samples with 128 sampling steps and token-channel classifier-free guidance with guidance weight $w=2.75$ (\cref{app:lm1b_cfg}), and report their mean GPT-2-Large perplexity\footnote{Following ELF~\citep{hu2026elf}, we do not report validation perplexity, since likelihood evaluation for flow-based models can require additional likelihood-specific training~\citep{ai2026joint}.}; the ground-truth row of \cref{tab:lm1b} is measured in the same way on LM1B text. The baseline results were obtained by the LangFlow authors after retraining the corresponding models on LM1B for 1M steps. Baseline parameter counts follow the same sampling-time, non-embedding rule. Plaid has only 0.5M embedding parameters, which we omit from its reported 109M count. LM1B training uses two stages and 660K steps in total: In the first stage, the encoder and token predictor are trained for 80K steps with a KL loss weighted by $1{\times}10^{-4}$. In the second stage, the denoiser is trained while the encoder and token predictor are kept fixed. We use AdamW with a global batch size of 224 and learning rate $3{\times}10^{-4}$ under a linear continuous-time noise schedule. Training is performed on a single NVIDIA RTX PRO 6000 GPU with bfloat16 precision and takes approximately 48 hours.

\paragraph{LM1B training efficiency.}
\Cref{tab:lm1b_training_time} provides the reported wall-clock training times as compute context rather than as a hardware-controlled efficiency comparison. Our model is trained in approximately 48 hours on a single NVIDIA RTX PRO 6000 GPU using bfloat16 precision. Table~4 of the LangFlow paper reports approximately 292 hours for LangFlow and the LM1B baselines retrained by its authors (Transformer, MDM, SEDD Absorb, and Duo), without a separate duration for each retrained model, and approximately 375 hours for Plaid. Because the hardware and training schedules differ, these figures should be interpreted as reported end-to-end training costs rather than direct throughput measurements.

\begin{table}[H]
\centering
\caption{Reported wall-clock training time for LM1B models.}
\label{tab:lm1b_training_time}
\setlength{\tabcolsep}{8pt}
{\small
\begin{tabular}{lc}
\toprule
Method & Training time \\
\midrule
LangFlow and retrained baselines$^\dagger$ & $\sim$292 h \\
Plaid & $\sim$375 h \\
\name{} (ours) & $\sim$48 h \\
\bottomrule
\end{tabular}
}
\end{table}

$^\dagger$ LangFlow reports a shared duration for LangFlow and its retrained baselines rather than per-model measurements; hardware and training schedules also differ across the reported runs.

\subsection{Sample Quality}
\label{app:lm1b_sample_quality}

\Cref{tab:lm1b_sample_quality} reports the generative perplexity and entropy across different numbers of neural function evaluations (NFEs). Both metrics are computed over 1,024 unconditional samples at each NFE.

\begin{table}[H]
\centering
\caption{Sample quality across sampling budgets. Lower Gen. PPL is better.}
\label{tab:lm1b_sample_quality}
\setlength{\tabcolsep}{9pt}
{\small
\begin{tabular}{lcc}
\toprule
NFE & Gen. PPL & Entropy \\
\midrule
128 & 75.5  & 4.21 \\
64  & 80.1  & 4.21 \\
32  & 93.0  & 4.21 \\
16  & 117.2 & 4.18 \\
8   & 168.8 & 4.02 \\
\midrule
\textit{Ground truth} & 40.4 & 4.32 \\
\bottomrule
\end{tabular}
}
\end{table}

\subsection{Inference Speed}
\label{app:lm1b_inference_speed}

\begin{figure}[t]
    \centering
    \includegraphics[width=0.6\linewidth]{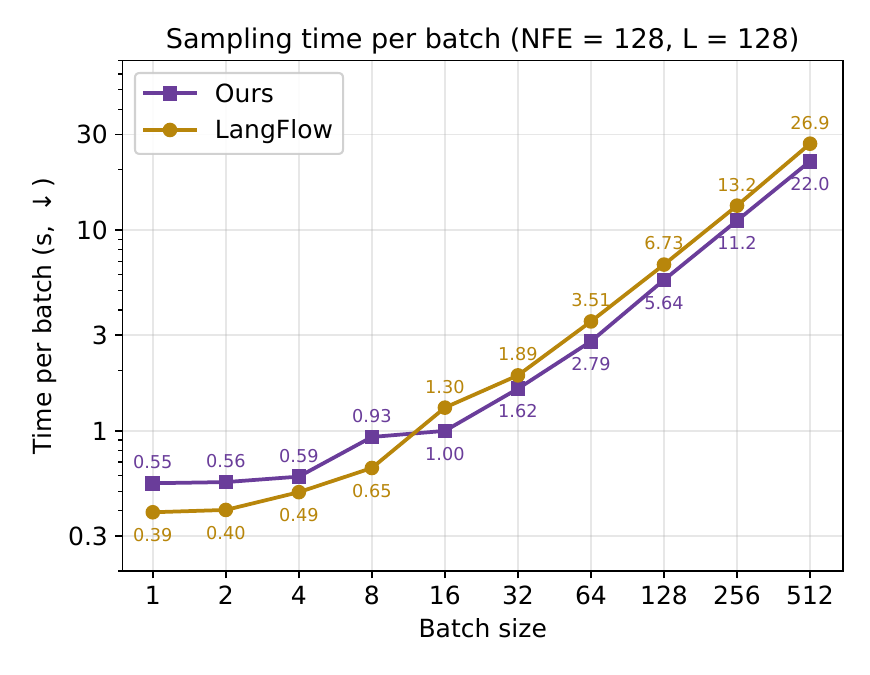}
    \caption{Wall-clock LM1B sampling time per batch at 128 NFEs and sequence length 128. \name{} is faster from batch size 16 onward, with the gap widening as the batch grows. Lower is better.}
    \label{fig:lm1b_speed}
\end{figure}

\Cref{fig:lm1b_speed} compares wall-clock sampling time per batch for \name{} and LangFlow under the same sampling budget and sequence length on a single NVIDIA RTX PRO 6000 GPU. At larger batch sizes, \name{} maintains a lower time per batch than LangFlow, suggesting favorable efficiency for batched parallel generation.

\subsection{Classifier-Free Guidance}
\label{app:lm1b_cfg}

\paragraph{Classifier-free guidance on the token channel.}
Although LM1B generation is unconditional at the sequence level, we use classifier-free guidance (CFG) to control how strongly the continuous denoiser follows its internal discrete token scaffold. At reverse step $t$, the denoiser takes the noisy continuous latent $x_t$, normalized time $t/T$, and token scaffold $k_t$, and predicts the clean latent, $d_\phi(x_t,k_t,t/T)$. During training, we first construct $k_t$ normally. With probability 0.2, we replace the entire row by the maximally noised channel at normalized time $t=1$, in which every position is an independently and uniformly sampled random token ID; we denote this uninformative channel by $\varnothing$. Because $\varnothing$ contains no information about the clean sequence, the same denoiser learns both
\begin{align}
d_\phi(x_t,k_t,t/T) &\approx \mathbb{E}[x_0\mid x_t,k_t],
&
d_\phi(x_t,\varnothing,t/T) &\approx \mathbb{E}[x_0\mid x_t].
\label{eq:token_channel_cfg_training}
\end{align}

\begin{figure}[t]
    \centering
    \includegraphics[width=0.6\linewidth]{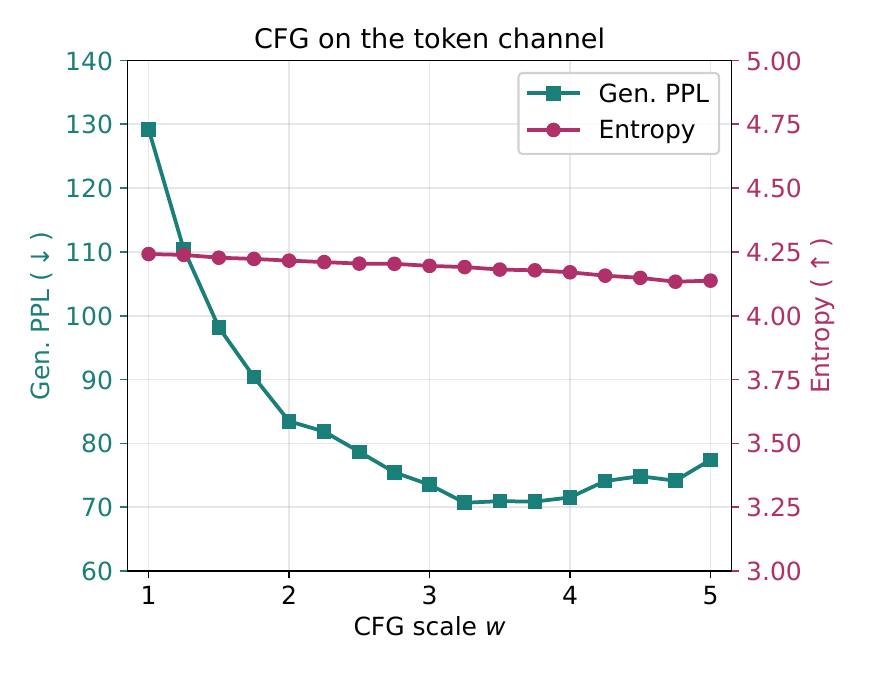}
    \caption{Effect of token-channel classifier-free guidance. We report generative perplexity (lower is better) and entropy (higher is better) as the guidance scale $w$ varies. The main LM1B result uses $w=2.75$; $w=1$ corresponds to sampling without guidance extrapolation.}
    \label{fig:lm1b_cfg}
\end{figure}

At each sampling step, we evaluate the denoiser twice at the same continuous state:
\begin{align}
    \hat{x}_{0,c} &= d_\phi(x_t,k_t,t/T),
    &
    \hat{x}_{0,u} &= d_\phi(x_t,\varnothing_t,t/T),
    \label{eq:token_channel_cfg_predictions}
\end{align}
where a fresh random-token sequence $\varnothing_t$ is sampled at every reverse step. We then extrapolate along the direction contributed by the token scaffold:
\begin{equation}
    \hat{x}_0^{\mathrm{CFG}}
    = \hat{x}_{0,u} + w\bigl(\hat{x}_{0,c}-\hat{x}_{0,u}\bigr).
    \label{eq:token_channel_cfg}
\end{equation}
Thus, $w=1$ recovers the ordinary scaffold-conditioned prediction, while $w>1$ increases reliance on the token channel. We use $\hat{x}_0^{\mathrm{CFG}}$ in the Euler update of the continuous state. The token predictor then decodes $\hat{x}_0^{\mathrm{CFG}}$ into a clean token sequence, which is corrupted according to the categorical forward kernel at the next noise level to produce the next scaffold $k_{t_{\mathrm{next}}}$. This guided continuous update and token refresh are repeated at every sampling step. \cref{fig:lm1b_cfg} compares generative perplexity and entropy as $w$ varies. Across the CFG sweep, proper guidance lowers generative perplexity while preserving entropy, indicating improved sample quality without reducing diversity.

\subsection{LM1B Generation Samples}
\label{app:lm1b_samples}

\Cref{tab:lm1b_samples} presents three unconditional generation examples from the LM1B experiment using 128 NFEs.

\begin{table}[H]
\centering
\caption{Unconditional samples generated by \name{} on LM1B.}
\label{tab:lm1b_samples}
\setlength{\tabcolsep}{6pt}
{\small
\begin{tabular}{p{0.12\linewidth}p{0.80\linewidth}}
\toprule
Example & Generated text \\
\midrule
Example 1 & first 21 points of the quarter because they rallied by 21 points .\textless{}\textbar{}endoftext\textbar{}\textgreater{}It is , after all , Obama 's favourite to win for president .\textless{}\textbar{}endoftext\textbar{}\textgreater{}Around 700,000 to 800,000 are required to pay their retails , including winter holidays , year-course , morning holidays , trips , and rigs taxes .\textless{}\textbar{}endoftext\textbar{}\textgreater{}Even with his lawyer , Dorothy Kinaitz , who has planned a separate appeal , he could soon appeal to the jury .\textless{}\textbar{}endoftext\textbar{}\textgreater{}Although Geoff Karlfeld 's Irish blood was redbed by judges Alec Dominan and Dr. McGerson , the American men \\
\midrule
Example 2 &  alcohol .\textless{}\textbar{}endoftext\textbar{}\textgreater{}This year , NDO has cut its profit forecast for the second quarter of 2009 .\textless{}\textbar{}endoftext\textbar{}\textgreater{}\" Yes , you are going to have a real team that is going to get the support and remain out of the race as to how very good they 're going through , \" McCain said .\textless{}\textbar{}endoftext\textbar{}\textgreater{}An aggressive Revolutionary Union militia , Tim Lappba , were killed in the fighting .\textless{}\textbar{}endoftext\textbar{}\textgreater{}The impact of the April quake was striking , with anticipated power managers warning from the city 's summer thunderstorms .\textless{}\textbar{}endoftext\textbar{}\textgreater{}The Blue Millions Rural Care is likely to be offered if the county fails to comply .\textless{}\textbar{}endoftext\textbar{}\textgreater{}Publishing professionals \\
\midrule
Example 3 &  .\textless{}\textbar{}endoftext\textbar{}\textgreater{}\" We have to work together to make sure that it is difficult to do all of the things that we don 't do in the future , \" he said .\textless{}\textbar{}endoftext\textbar{}\textgreater{}Meanwhile , a couple of elections in the northern port city of Yivazoum sent a complaint saying his support for Mr Tsvangirai and Mr Mugabe to a disastrous bid for a second term .\textless{}\textbar{}endoftext\textbar{}\textgreater{}Bill Greenberg , a Republican on the Finance Committee , said that if Congress was for delay , the State program linked to Bush 's overhaul of the financial institutions was not upstate to slip out the majority .\textless{}\textbar{}endoftext\textbar{}\textgreater{}Undle Sarben \\
\bottomrule
\end{tabular}
}
\end{table}

\subsection{Latent Sequence Length}
\label{app:m_ablation}

The latent sequence length $M$ controls the capacity of the continuous channel: larger $M$ provides more continuous tokens for the continuous denoiser to operate on at the cost of additional compute. \cref{tab:ablation_m} reports Sudoku accuracy on the Easy and Hard splits as $M$ varies in $\{1, 2, 4, 8, 16\}$, with all other hyperparameters held fixed at the structured-task values in \cref{app:train}. The table suggests that increasing $M$ improves accuracy up to a specific capacity ($M = 4$), hinting at an optimal level of latent expressivity before performance degradation.

\begin{table}[h]
\centering
\caption{Ablation on latent sequence length $M$ (Sudoku, accuracy \%$\uparrow$). All other hyperparameters follow the main configuration in \cref{app:train}; $M=1$ is the setting used throughout the Sudoku and Countdown experiments, while LM1B defaults to $M=128$, with one latent per token.. For a fair comparison of early-to-mid stage optimization efficiency, we report accuracy at the same checkpoint (560 epochs). }
\label{tab:ablation_m}
\setlength{\tabcolsep}{10pt}
\begin{tabular}{rcc}
\toprule
$M$ & Easy & Hard \\
\midrule
1  & 81.73   & 55.81   \\
2  & 91.59 & 67.34 \\
4  & \textbf{93.75} & \textbf{71.11} \\
8  & 71.82 & 45.52 \\
16 & 78.80 & 51.89 \\
\bottomrule
\end{tabular}
\end{table}

\subsection{Wall-Clock Inference Time}
\label{app:inference_time}

\begin{figure}[t]
    \centering
    \includegraphics[width=0.6\linewidth]{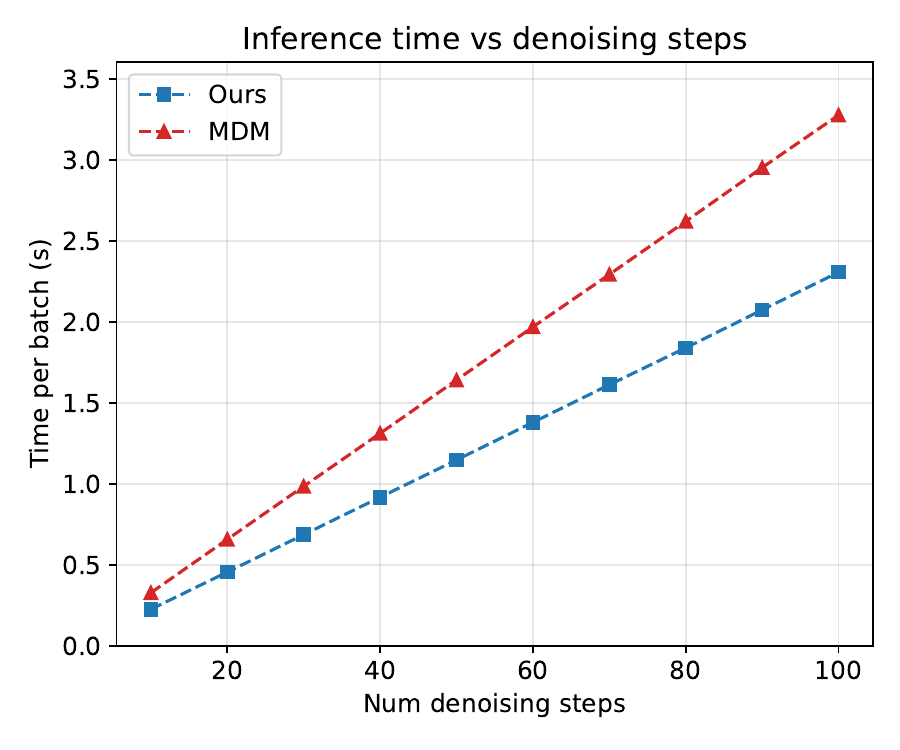}
    \caption{\textbf{Wall-clock inference time versus denoising steps.} Runtime per generated batch scales approximately linearly for both methods. Across tested step counts, \name{} is consistently faster, reflecting its efficient DiT implementation.}
    \label{fig:denoising_time}
\end{figure}

The step-efficiency results in \cref{fig:repr_trajectory_sub2} compare final accuracy as the number of denoising steps is reduced. Since each reverse step may have a different computational cost, we additionally report wall-clock inference time in \cref{fig:denoising_time}. The measurement uses the same evaluation implementation and batch setting as the denoising-step ablation. 

\cref{fig:denoising_time} shows that \name{} is consistently faster than MDM at the same number of steps despite using an additional continuous state. At 100 denoising steps, \name{} takes roughly $2.3$ seconds per batch, compared with about $3.3$ seconds for MDM. Together with \cref{fig:repr_trajectory_sub2}, this indicates that \name{} improves solution quality without increasing measured sampling latency in our implementation.

\subsection{Why the Uniform Kernel Suits the Hierarchy}
\label{app:uniform}

This section expands the analysis of the uniform--absorbing gap observed in \cref{tab:ablation_ordering_noise}. Two properties of the hierarchy explain it. The first is that the discrete state acts as a conditioning scaffold for the latent denoiser rather than as a ledger of commitments. A uniformly corrupted sequence is a noisy but complete hypothesis in which every position carries a concrete token, whereas absorbing corruption replaces positions with \textsc{[mask]} symbols that carry no information. In our hierarchy, the ``information void'' that motivates continuous augmentation in CADD~\citep{zheng2025continuously} instead impoverishes the scaffold itself. The second is that tokens are read out from the latent anew at every step, so the uniform kernel keeps every position revisable throughout generation. Adaptive unmasking under the absorbing kernel instead freezes commitments one by one, which is costly in constraint-dense tasks where a single early error invalidates the whole solution.

\section{Extended Comparison with Hybrid Discrete-Continuous Diffusion}
\label{app:comparison}

This section expands the comparison of \cref{sec:related} with three representative hybrid models that attach the continuous state to a discrete chain in different ways, namely a global latent drawn once (VMD), per-token hints (CADD), and a parallel embedding chain (CCDD). \cref{tab:comparison} summarizes the structural differences, and the paragraphs below give the details.

\begin{table*}[t]
\centering
\small
\setlength{\tabcolsep}{4pt}  
\renewcommand{\arraystretch}{1.2}
\begin{tabular}{@{}lcccc@{}}
\toprule
Method & \makecell{Continuous\\variable}
       & \makecell{Update of the\\continuous variable}
       & \makecell{Input of the\\token predictor}
       & \makecell{Decoded token\\can change later} \\
\midrule
MDM            & none                    & ---                            & $k_t$            & no \\
VMD            & global latent           & not updated (sampled once)     & $k_t,\,z$        & no \\
CADD           & noised token embeddings & no learned update              & $k_t,\,x_t$      & no \\
CCDD           & frozen token embeddings & learned denoiser               & $k_t,\,x_t$      & no \\
\midrule
\name{} (ours) & global latent           & learned denoiser               & $\hat{x}_0$ only & yes \\
\bottomrule
\end{tabular}
\caption{Structural comparison with hybrid discrete--continuous diffusion models. MDM~\citep{sahoo2024simple}, VMD~\citep{zhang2025variational}, CADD~\citep{zheng2025continuously} and CCDD~\citep{zhou2025coevolutionary} use the absorbing kernel, under which a token stays fixed once it is unmasked. For \name{}, $\hat{x}_0$ is the denoiser's clean-latent estimate.}
\label{tab:comparison}
\end{table*}

\subsection{VMD}
\label{app:cmp_vmd}

Variational Masked Diffusion (VMD)~\citep{zhang2025variational} extends masked diffusion with a single global latent $z$ drawn once from a Gaussian prior and held fixed for all generation steps, and derives a variational bound that jointly trains an encoder of $z$ and the masked diffusion model conditioned on it. The latent supplies static global context that cannot adapt as the partially decoded sequence changes, and the discrete chain remains the generative process. \name{} keeps VMD's learned variational latent but turns it into a denoising trajectory of its own, refined at every step under feedback from the token scaffold.

\subsection{CADD}
\label{app:cmp_cadd}

CADD~\citep{zheng2025continuously} keeps absorbing masked diffusion as the primary process and pairs each token with a continuous hint. The clean hint is the token's row in the model's own learnable embedding table $w_\theta$, and it is corrupted by Gaussian noise only from the moment its token is masked. The hint has no learned denoiser. A position revealed at a step copies the embedding of its sampled token, and a position that stays masked takes a Gaussian posterior step towards an estimate computed from the predicted token distribution $\pi_\theta(\cdot\mid k_t,x_t)$, by default the embedding of its most likely token. The hint therefore stores the model's own earlier predictions and feeds them back to the token predictor as an extra input, so it carries no information that the token predictor did not already produce. Two consequences follow. The discrete part of their posterior does not depend on the hint (their Proposition~2), so the discrete chain alone decides which positions unmask and when. And because the hint is corrupted only once its token is masked, the construction requires the absorbing kernel and cannot use the uniform one.
In \name{} the dependence runs the other way. The latent has a learned denoiser of its own, and the tokens are read out of it.

\subsection{CCDD}
\label{app:cmp_ccdd}

CCDD~\citep{zhou2025coevolutionary} defines a joint process over tokens and \emph{frozen} embeddings of the clean sequence from a pretrained text encoder (one contextualized vector per position, e.g., from Qwen3-Embedding). As in our \cref{eq:forward}, the forward process corrupts the tokens and the embeddings independently. The reverse process is symmetric. A single backbone with two parallel heads predicts both clean states from the joint input, and each modality is then updated by its own standard rule, so the two channels interact only through the shared network inputs.
Its objective is the weighted sum of the two single-channel diffusion losses. The token head reads the joint hidden state, so the token update sees $k_t$ directly, and under the absorbing kernel an unmasked token stays fixed. \name{} has no separate token head, since its tokens come from the latent alone. Our experiments use this design on our backbone and learned latent rather than frozen per-token embeddings (\cref{app:train}).

\subsection{Summary}
\label{app:hybrid_summary}

\name{} differs from all three hybrids in the last two columns of \cref{tab:comparison}. Its token predictor reads only the latent, so every token is re-read from the latent at each step instead of being carried over once decoded. This suits the uniform kernel, which in our experiments works clearly better than the absorbing one (\cref{tab:ablation_ordering_noise}).

\section{Limitations}
\label{app:limitations}

\name{} jointly optimizes an encoder, a continuous denoiser, and a token predictor, which makes each training step more expensive than in a purely discrete masked diffusion baseline. This overhead is confined to training: the encoder is discarded at sampling time, and inference remains faster than the baseline at matched step counts (\cref{app:inference_time}). Our concatenation-based conditioning on $k_t$ is simple and effective, though more structured fusion mechanisms may further improve sample efficiency. Finally, our experiments target structured reasoning and planning benchmarks at moderate scale, where correctness can be verified exactly and controlled comparisons fit an academic compute budget. Since all modules are standard DiT blocks and the latent channel adds only $M{=}1$ token, the design carries no scale-specific components; scaling \name{} to larger pretrained backbones is therefore mainly a question of compute, which exceeds the resources typically available in academia, and we consider it the most promising direction for future work.

\end{document}